\documentclass[acmlarge]{acmart}

\AtBeginDocument{\providecommand\BibTeX{{\normalfont B\kern-0.5em{\scshape i\kern-0.25em b}\kern-0.8em\TeX}}}

\setcopyright{acmlicensed}
\acmJournal{IMWUT}
\acmYear{2022} \acmVolume{6} \acmNumber{3} \acmArticle{153} \acmMonth{9} \acmPrice{15.00}\acmDOI{10.1145/3550291}

\acmSubmissionID{2991}

\usepackage{xspace}
\usepackage{balance}
\usepackage{amsmath}
\usepackage{multirow}
\usepackage[textfont={normal}]{caption} 
\usepackage[labelformat = parens, labelsep = space, textfont=small]{subcaption}

\usepackage{enumitem}

\newcommand{\sysname}{\textsc{LitAR}\xspace}

\newcommand{\shc}{spherical harmonics coefficients\xspace}
\newcommand{\SHc}{SH coefficients\xspace}

\newcommand{\uspc}{unit sphere-based point cloud\xspace}

\newcommand{\Xihe}{Xihe\xspace}
\newcommand{\reconstruction}{two-field lighting reconstruction\xspace}
\newcommand{\extrapolate}{anchor extrapolation\xspace}
\newcommand{\dpcp}{multi-resolution projection\xspace}
\newcommand{\lrs}{lighting reconstruction session\xspace}

\definecolor{pro_green}{rgb}{0.0, 0.66, 0.47}

\newcommand{\1}{{\em (i)}}
\newcommand{\2}{{\em (ii)}}
\newcommand{\3}{{\em (iii)}}

\setlist[itemize]{leftmargin=.12in}

\begin{document}

\title{\sysname: Visually Coherent Lighting for Mobile Augmented Reality}

\author{Yiqin Zhao}
\orcid{0000-0003-1044-4732}
\affiliation{\institution{Worcester Polytechnic Institute}
 \streetaddress{100 Institute Road}
 \city{Worcester}
 \state{MA}
 \country{USA}
 }
\email{yzhao11@wpi.edu}

\author{Chongyang Ma}
\orcid{0000-0002-8243-9513}
\affiliation{\institution{Kuaishou Technology}
 \streetaddress{6 Shangdi West Road Haidian}
 \city{Beijing}
 \state{Beijing}
 \country{China}
 }
\email{chongyangm@gmail.com}

\author{Haibin Huang}
\orcid{0000-0002-7787-6428}
\affiliation{\institution{Kuaishou Technology}
 \streetaddress{6 Shangdi West Road Haidian}
 \city{Beijing}
 \state{Beijing}
 \country{China}
 }
\email{jackiehuanghaibin@gmail.com}

\author{Tian Guo}
\orcid{0000-0003-0060-2266}
\affiliation{\institution{Worcester Polytechnic Institute}
 \streetaddress{100 Institute Road}
 \city{Worcester}
 \state{MA}
 \country{USA}
 }
\email{tian@wpi.edu}

\renewcommand{\shortauthors}{Zhao et al.}

\begin{abstract}

An accurate understanding of omnidirectional environment lighting is crucial for high-quality virtual object rendering in mobile augmented reality (AR).
In particular, to support reflective rendering, existing methods have leveraged deep learning models to estimate or have used physical light probes to capture physical lighting, typically represented in the form of an environment map.
However, these methods often fail to provide visually coherent details or require additional setups. 
For example, the commercial framework ARKit uses a convolutional neural network that can generate realistic environment maps; however the corresponding reflective rendering might not match the physical environments.
In this work, we present the design and implementation of a lighting reconstruction framework called \sysname that enables realistic and visually-coherent rendering.
\sysname addresses several challenges of supporting lighting information for mobile AR. 

First, to address the spatial variance problem, \sysname uses \reconstruction to divide the lighting reconstruction task into the spatial variance-aware near-field reconstruction and the directional-aware far-field reconstruction.
The corresponding environment map allows reflective rendering with correct color tones.
Second, \sysname uses two noise-tolerant data capturing policies to ensure data quality, namely guided bootstrapped movement and motion-based automatic capturing.
Third, to handle the mismatch between the mobile computation capability and the high computation requirement of lighting reconstruction, \sysname employs two novel real-time environment map rendering techniques called \dpcp and anchor extrapolation.
These two techniques effectively remove the need of time-consuming mesh reconstruction while maintaining visual quality.
Lastly, \sysname provides several knobs to facilitate mobile AR application developers making quality and performance trade-offs in lighting reconstruction.
We evaluated the performance of \sysname using a small-scale testbed experiment and a controlled simulation. 
Our testbed-based evaluation shows that \sysname achieves more visually coherent rendering effects than ARKit. 
Our design of \dpcp significantly reduces the time of point cloud projection from about 3 seconds to 14.6 milliseconds.
Our simulation shows that \sysname, on average, achieves up to 44.1\% higher PSNR value than a recent work Xihe on two complex objects with physically-based materials.

\end{abstract}

\ccsdesc[500]{Computing methodologies~Mixed / augmented reality}
\ccsdesc[500]{Human-centered computing~Ubiquitous and mobile computing systems and tools}
\ccsdesc[500]{Computer systems organization~Distributed architectures}

\keywords{mobile augmented reality; lighting estimation; 3D vision}

\begin{teaserfigure}
    \includegraphics[width=1\linewidth]{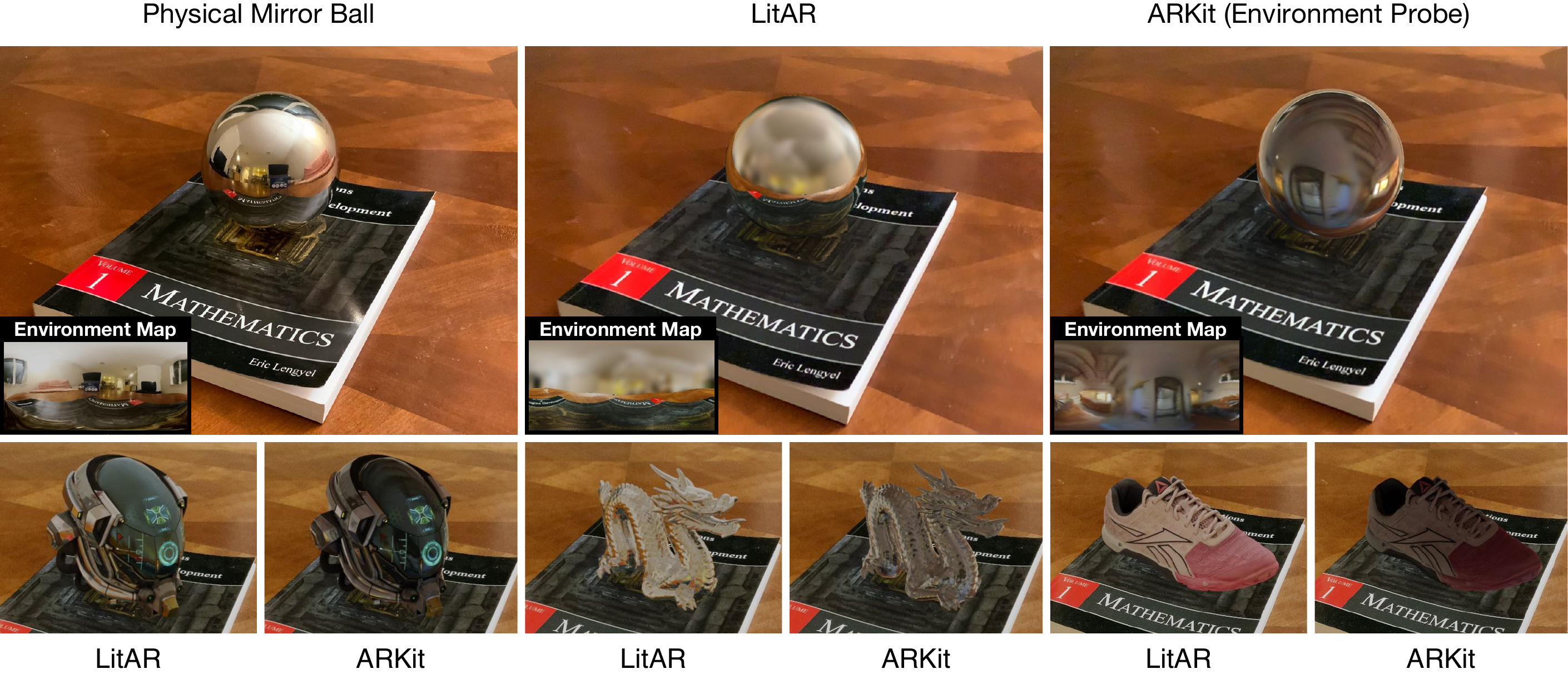}
    \caption{
        Rendered objects with \sysname compared to a physical mirror ball and with ARKit~\cite{arkit}.
        \textnormal{
        \emph{Row 1:} \sysname produces visually-coherent reflective rendering while ARKit fails; the bottom part of the ARKit-rendered ball does not faithfully reflect the book cover. The top part of the \sysname-rendered ball has an intentionally gradient blurring effect for quality-performance trade-offs (see \S\ref{subsec:two_field_lighting_recon}).
        \emph{Row 2:} \sysname achieves more realistic and visually coherent rendering effects than ARKit for objects with different materials. 
        }
    }
    \label{fig:teaser}
\end{teaserfigure}

\maketitle

\section{Introduction}
\label{sec:introduction}

Mobile augmented reality (AR) has attracted increasing interest from academia and industry to better engage users by allowing seamless integration of physical and virtual environments~\cite{Google_for_Education_undated-rj,Inter_IKEA_Systems_B_V2017-iv,Andrews2019-qh}.
The current mobile AR ecosystem is infused with new hardware development~\cite{ipad-lidar}, improved frameworks~\cite{arkit}, advancing vision and graphics algorithms~\cite{pointar_eccv2020}, as well as end-user facing applications ranging from e-commerce ones to educational ones~\cite{Amazon_undated-vp}.

Given the interactive nature of AR applications, users often prefer virtual objects of high visual quality.
The rendered virtual objects should look realistic and feel like they belong to the physical surroundings, a property commonly referred to as \emph{visual coherence}. 
For example, virtual sunglasses that are overlaid on a user's face should look like physical sunglasses (realistic) and reflect the correct physical environment (visually coherent). 
To achieve both realistic and visually coherent rendering, mobile AR applications require access to an accurate representation of omnidirectional environment lighting (often represented as \emph{environment map} for image-based lighting) at the user-specified rendering position~\cite{pointar_eccv2020,prakash2019gleam}.

However, obtaining a high-quality environment map for mobile AR has to overcome several key challenges.
First, the inherently spatial variation of indoor environment lighting makes the environment map at the \emph{observation position}---which can be more easily reconstructed with more camera observations---a poor approximation of the \emph{rendering position} environment map.
This challenge was demonstrated in prior work~\cite{Garon2019,pointar_eccv2020} and our motivating example (see Figure~\ref{fig:motivation}). 
Second, the natural user mobility of mobile AR usage can induce noise to necessary data (such as 6DoF tracking and RGB image data) for lighting estimation. 
For example, we observe that the tracking data provided by ARKit can show that consecutive camera frames are misaligned, although they represent the same physical space.
Third, mobile devices can have heterogeneous sensing capability, e.g., in terms of cameras' field-of-view (FoV) or their depth-sensing ability, which makes it necessary to consider auxiliary components (such as depth estimation~\cite{zhang-indepth,Alhashim2018,Ranftl2020}) for obtaining accurate lighting.  
Fourth, mobile devices have relatively limited resources compared to their desktop counterpart, while the interactivity nature often requires 30 fps rendering.
The computational limitation makes it challenging to directly use computational-intensive models designed to run on powerful GPU servers~\cite{Song2019}, and motivates minimal usage or optimization of time-consuming operations (such as point cloud registration~\cite{besl1992method} and mesh reconstruction~\cite{bernardini-ball-pivoting}).

In this work, we investigate the problem of \emph{providing high-quality lighting information for mobile AR} by addressing the above four key challenges.  
Our key goal is to support realistic and visually coherent rendering of virtual objects with various geometries and materials. 
Figure~\ref{fig:teaser} compares the visual effect of virtual objects rendered with lighting information obtained with our proposed system called \sysname and ARKit.
We show that \sysname achieves high-quality rendering with structurally similar reflections with the physical object and more visually coherent reflection than objects rendered with ARKit.

\sysname involves a novel technique called \emph{\reconstruction} and several complementary components that work together to deliver a high-quality environment map with low-performance impact.
We design the \reconstruction with the insight of dividing the camera observations into two types, i.e., the near-field and far-field observations, to speed up lighting reconstruction while maintaining the visual quality. 
This technique shares a similar spirit to the well-known screen space reflection~\cite{mcguire2014efficient} and is tailored to mobile AR by fully exploring user mobility. 
Specifically, \sysname generates a multi-view dense point cloud to represent \emph{near-field} observations, corresponding to the portion of the environment map that receives more accurate and higher confidence depth information surrounding the rendering position.
This design helps produce geometrically accurate lighting transformation between the observation and rendering positions, thus supporting key rendering features like reflections and providing visually coherent results.
Furthermore, \sysname leverages far-field observations to handle the anisotropic lighting property by reconstructing sparse point clouds to reduce visual errors. 

On top of the \reconstruction, we incorporate two noise-tolerant data capturing policies, i.e., guided bootstrapped movement and motion-based automatic capturing, to improve the data quality.
The \emph{guided bootstrapped movement} policy directs the camera views to capture required near-field and far-field observations efficiently.
This policy also brings other benefits, such as enlarged FoVs and reduced user movement, for reconstructing high-quality lighting.
It is worth noting that \sysname can leverage new observations, e.g., device orientation and user movement, to improve the quality of the environment map progressively.
The \emph{motion-based automatic capturing} policy leverages multi-sensory information to capture spatially and temporally new observations. 
Moreover, we propose two performance optimizations that significantly reduce the time reconstructing the final environment map from the intermediate 3D point clouds.
The first optimization uses a lightweight \dpcp instead of the traditional expensive mesh reconstruction to generate the near-field portion.
The second optimization uses a unit sphere-based approach called \emph{anchor extrapolation} to generate gradient coloring and blurring effect of the far-field portion. 
Lastly, \sysname supports reconstruction quality and time trade-offs to account for dynamic lighting conditions. 
By default, \sysname provides three quality presets for mobile AR developers.

We implement \sysname as an edge-assisted framework that consists of about 2.2K lines of code running on both the mobile device and the edge server. 
Specifically, the client-side component works with various sensors, including color, depth, and motion sensors, to capture near-field and far-field observations. 
The resulting data is encoded and sent to the edge server to generate a fixed-size unit sphere point cloud and multi-view dense point clouds with good alignment and visual pixel continuity. 
Mobile AR applications built with Unity ARFoundation can directly use \sysname to render realistic virtual objects. We also implement a reference iOS AR application for our testbed-based evaluation.  
To evaluate \sysname in a controlled environment, we develop a simulator based on Unreal Engine that exposes multiple knobs for controlling physical factors such as camera location and FoV while providing ground truth lighting information. 
The testbed-based system evaluation shows that \sysname achieves more visually coherent rendering results and higher PSNR/SSIM values than ARKit for three real-world scenes.
The end-to-end latency measurements show that \sysname can generate about 22 environment maps per second, effectively supporting 22 fps which is sufficient for most mobile AR applications~\cite{Xu2021-sn, Yi2020-na}.
Our simulator-based evaluations include one realistic indoor scene and six virtual objects of different shapes and materials.
We evaluate the performance of \sysname with various observation-rendering position pairs. 
We show that it achieves 36.7\% and 17.1\% higher rendering PSNR compared to a recent deep learning-based lighting approach~\cite{xihe_mobisys2021} and the environment lighting captured by 360$^{\circ}$ cameras at the observation position, respectively. 

Related work on generating spatially-varying lighting includes classical physical probe-based techniques~\cite{debevec2006image,debevec2008rendering,prakash2019gleam} and learning-based solutions~\cite{Garon2019,pointar_eccv2020,Song2019,Somanath2020-of}.
Physical probe-based techniques often produce high-quality environment maps but have more constrained usage scenarios since they require additional setup~\cite{Somanath2020-of}.
On the other hand, the applicability of learning-based solutions is often limited by the access to extensive training datasets, e.g., Matterport3D~\cite{chang2017matterport3d}, and their suitability to run on heterogeneous mobile devices~\cite{Song2019}.
Another side effect is the difficulty of conducting comprehensive comparisons due to the lack of publicly available source code and benchmark dataset~\cite{Somanath2020-of}.
In this work, we are interested in designing a mobile-specific lighting framework that circumvents the above-mentioned limitations by considering mobile characteristics from the outset. 
Compared to a recent method by Somanath et al.~\cite{Somanath2020-of} that generates HDR environment maps using a neural network based on adversarial training, \sysname has the advantage of simplicity yet achieving good visual coherence. 

In summary, we make the following key contributions:
\begin{itemize}[leftmargin=.12in,topsep=4pt]
    \item We design a novel technique called \reconstruction, which generates high-quality environment maps from mobile cameras with limited FoV.
    Each environment map consists of near-field and far-field portions, separately constructed from near-field and far-field observations.
    The resulting environment map captures spatial and directional variances and is suitable for reflective rendering.
    \item We develop several complementary approaches to handle mobility-induced noise, limited mobile sensing capabilities, and the computation intensity that naturally arises during the lighting reconstruction process. 
    For example, our \dpcp and anchor extrapolation techniques efficiently project the intermediate 3D point clouds to the final 2D environment maps. 
    These techniques ensure high data input quality, good usability, and low reconstruction time.
    \item We implement the entire framework as an edge-assisted system called \sysname and develop a simulator based on Unreal Engine for evaluation purposes. 
    The system implementation provides a platform to compare \sysname to the commercial framework ARKit.
    The simulator facilitates controlled experiments and allows easy comparisons between lighting techniques and ground truth lighting.
    Our source code and related artifacts are available at \url{https://github.com/cake-lab/LitAR} to encourage follow-up research.
\item We evaluate the performance of \sysname on a small-scale testbed using the simulator. The testbed-based system evaluation shows that \sysname (at all three quality presets) outperforms ARKit in three real-world indoor scenes. \sysname also delivers environment maps at 22 fps or even higher, depending on the quality settings. The simulation-based evaluation shows that \sysname can achieve up to 36.7\% higher PSNR values on objects with various geometries and materials than a recent lighting framework~\cite{xihe_mobisys2021}.
\end{itemize}

\section{Background: Lighting For Mobile AR} 

Obtaining lighting information is a classic problem in computer vision and computer graphics~\cite{Einabadi2021-vq}.
Access to accurate environment lighting information is crucial to many applications related to photorealistic rendering and image manipulation, such as 3D object composition~\cite{Gardner2019-gc} and portrait relighting~\cite{Pandey2021-wn}.

As the capability of mobile devices improves and AR re-emerges in user-facing applications~\cite{Amazon_undated-vp, Inter_IKEA_Systems_B_V2017-iv}, obtaining environment lighting for photorealistic rendering has gained increasing interests in various research communities~\cite{ChengSCDZ18_graph,prakash2019gleam,pointar_eccv2020,Somanath2020-of,xihe_mobisys2021}.
In addition to these mobile-specific lighting techniques, researchers have investigated image-based lighting~\cite{debevec2006image,corsini2008stereo,karis2013real}, generating environment lighting representations from camera videos~\cite{havran2005interactive,unger2013temporally,grosch2007consistent}, and assisted lighting reconstruction with physical probes~\cite{debevec2008rendering}, object cues~\cite{Sun2019-rw}, or scene geometry~\cite{Maier2017-qp,Azinovic2019-nz}.
Recent work is mostly deep learning-based and can be broadly categorized based on the output: estimating low-frequency lighting~\cite{Garon2019,xihe_mobisys2021} or high-quality lighting~\cite{Gardner2017,Song2019}. 

This paper aims to provide lighting support for mobile augmented reality (AR), an emerging application that augments the real world by overlaying with virtual objects in indoor scenes.
Our work will develop a non-learning-based approach by leveraging \emph{lighting reconstruction} (\S\ref{sec:lighting_reconstruction_premier}) to address the following two issues:
\1 the typical limitations of deep learning based methods such as training data availability and inference performance on heterogeneous mobile resources;
\2 the underexploited mobile characteristics.
Note that techniques commonly used for creating realistic virtual worlds, such are screen space rendering~\cite{mcguire2014efficient} that only ray traces what is being presented on the screen, can fall short in delivering the visual coherence required by AR.
This shortcoming is mainly due to the key difference in perceivable lighting impact on virtual objects; unlike in fully immersive environments, AR users can observe environment lighting outside screen space and potentially perceive incoherent visual artifacts.

Furthermore, we will focus on developing lighting understanding techniques for image-based rendering~\cite{debevec2006image}, e.g., representing lighting in the form of an \emph{environment map} to render virtual objects with different materials.
Our key goal is to generate high-quality environment maps for visually coherent rendering for mobile AR while keeping the time cost low.  
Specifically, we aim to reduce the overall time to obtain the final environment map and the component-wise time for intermediate outputs (such as point cloud projection).

\begin{figure}
\centering
    \includegraphics[width=\linewidth]{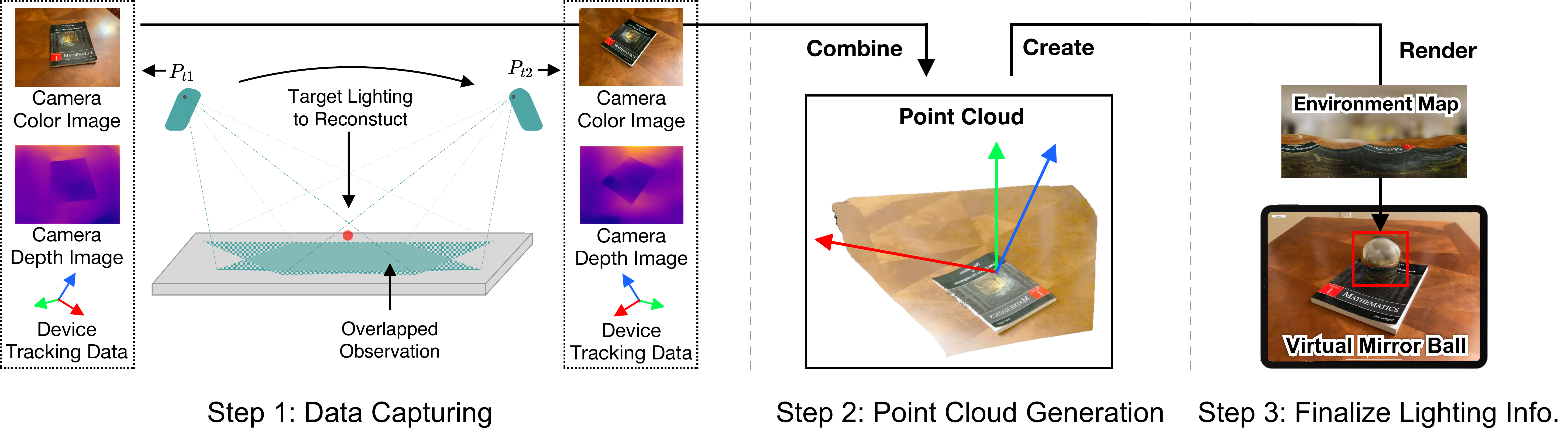}
    \caption{A simplified workflow of mobile AR lighting reconstruction.
        \textnormal{
            In this example, the mobile device starts at position $P_{t1}$ and points toward the rendering position on the table specified by the mobile user.
            Data such as color and depth images, as well as device tracking data, will be captured.
            The device will further capture more data (e.g., color/depth images from a different viewing direction) at position $P_{t2}$, which naturally overlaps with the data captured at $P_{t1}$.
            The data is maintained in a fixed-size buffer and will be aggregated to generate a global multi-view point cloud. 
            We will update the point cloud as new data arrives and use it to generate the final environment map for virtual object rendering.
        }
    }
    \label{fig:ar_workflow}
\end{figure}

\subsection{Lighting Reconstruction Primer} 
\label{sec:lighting_reconstruction_premier}

In this section, we present the critical information of \emph{multi-view lighting reconstruction} (refer to Figure~\ref{fig:ar_workflow}), which serves as the basis for understanding our proposed lighting reconstruction framework \sysname.
At a high level, we define \emph{lighting reconstruction} as a task similar to multi-view 3D reconstruction~\cite{openmvs2020,moulon2016openmvg} with the key difference in the reconstruction target. 
Our key insight is that by leveraging multiple captures of the physical environment that are often required by commercial AR frameworks~\cite{arkit,arcore}, we can use typical techniques used by 3D reconstruction to understand the environment lighting.
Conceptually simple, we have to address several challenges specific to the mobile AR environment (detailed in \S\ref{sec:motivation}).
Next, we describe the general procedure for the lighting reconstruction task.

\subsubsection{Step 1: Capture Environment Data.}
Several different types of data, e.g., color images and depth information, are needed in lighting reconstruction.
The common way to obtain these required data is to leverage a modern mobile device with a reasonable camera directly and to have the mobile AR users move the mobile device to scan the surroundings manually. 
The resulting captured data is often in the format of LDR or HDR images, which can then be used to reconstruct the environment's appearance and geometry. 
To improve the reconstruction quality and performance, one can also resort to additional setups such as using a physical chrome ball~\cite{prakash2019gleam,debevec2006image} or additional mobile sensors such as depth sensors~\cite{Huawei_undated-ca,ipad-lidar} and accelerometers.
Ambient light sensors can also be used to observe the ambient color, which helps match the object's color tone with the environment's lighting.
In this work, we focus on mobile devices that can capture color and depth images and provide device tracking data, e.g., a LiDAR-equipped iPad Pro.
Data will be captured from different viewing positions and used in the next step for generating a multi-view point cloud.

\subsubsection{Step 2: Generate a Point Cloud.}
Similar to other 3D reconstruction tasks~\cite{openmvs2020,moulon2016openmvg}, we convert the camera color, depth images, and device tracking data, into a point cloud-based representation in the world space.
The point cloud data structure allows us to combine the subsequent view data more efficiently than directly stitching 2D images.
Two practical issues often need to be addressed. 
First, real-world device tracking data can be noisy; one way to handle this issue is to use point cloud registration techniques such as the iterative closest point registration~\cite{besl1992method} to align the points. 
Second, some points might not have accurate depth information; to ensure the reconstruction quality, only points with high depth confidence values, which measure the accuracy of depth data, should be used. 
Note that we will update the point cloud based on newer data; conceptually, such an update helps deal with both spatial and temporal variance by initializing/overriding points in the 3D space at different times.

\subsubsection{Step 3: Finalize Environment Lighting.}
The generated multi-view point cloud consists of rich environment information and is equivalent to having an enlarged virtual camera FoV at the rendering position. 
It is worth noting that enlarging the camera FoV at the observation position can also increase the camera observation coverage, though less effective than multi-view enlargement. 
To directly use modern rendering engines to support realistic rendering, we convert the point cloud to lighting formats, such as spherical image format or environment map.
For example, one can project the point cloud into a 2D environment map that captures the omnidirectional environment lighting.

\section{Motivation and Challenges}
\label{sec:motivation}

\begin{figure}[t]
    \begin{subfigure}[b]{0.47\columnwidth}
        \centering
        \includegraphics[width=\linewidth]{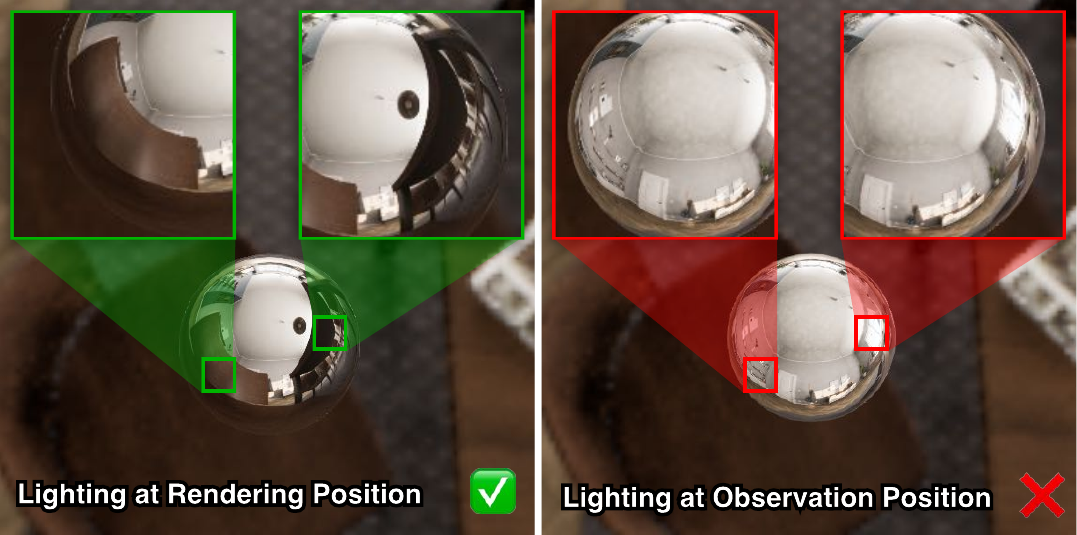}
        \caption{Spatial lighting variance}
        \label{fig:motivation:lighting_variance}
    \end{subfigure}\quad
    \begin{subfigure}[b]{0.47\columnwidth}
        \centering
        \includegraphics[width=\linewidth]{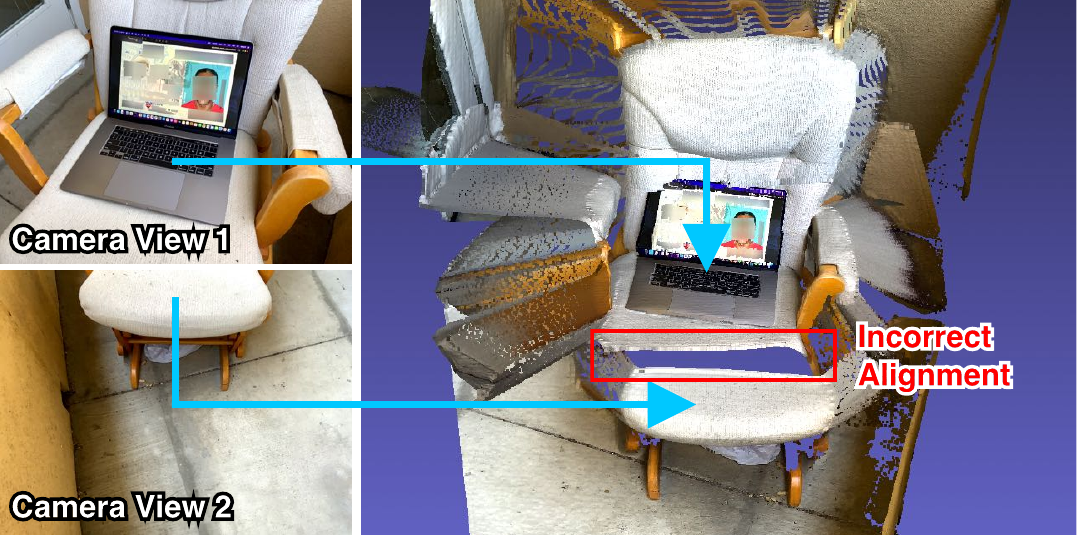}
        \caption{Mobility-induced noise}
        \label{fig:motivation:noise}
    \end{subfigure}
\caption{
        Motivating examples.
        \textnormal{
            (a) The same mirror ball object rendered with environment maps at the AR mobile device's position and the mirror ball placement position can have very different visual appearances. 
            (b) The mobility-induced tracking noise leads to a misaligned point cloud generated from two consecutive frames.
        }
    }
    \label{fig:motivation}
\end{figure}

\subsection{Spatial and Temporal Variance}
Indoor lighting can be both spatially and temporally varying~\cite{srinivasan20lighthouse,pointar_eccv2020,Garon2019}.
Rendering virtual objects using lighting information from locations other than the \emph{rendering position} may lead to potential visual degradation. 
To demonstrate the impact of such variances on the rendering effect, we compare a virtual object rendered with the lighting information at the rendering position and at the \emph{observation position} in Figure~\ref{fig:motivation:lighting_variance}.
We can see that the mirror ball on the right does not have the desired visual appearance, i.e., neither of the zoomed-in views contains the correct reflections of the chair and the table.
Thus, it is crucial to account for spatial variance when designing the reconstruction framework.
Intuitively, lighting can change temporally even at the same rendering position. 
In this work, we handle the temporal variance by periodically reconstructing lighting.

\subsection{Mobility-induced Noise}
The inherent user mobility can naturally introduce noise to data required by the lighting reconstruction pipeline.
For example, when a mobile user engages with an AR application, the user might move around, introducing measurement errors into the 6DoF tracking data or leading to blurred RGB images. 
Other factors, such as the mobile camera's position relative to the rendering position, can also impact the quality of camera observations.
Both tracking and camera data are commonly used for lighting estimation~\cite{pointar_eccv2020,xihe_mobisys2021}.
Figure~\ref{fig:motivation:noise} shows that the point cloud can have incorrect alignment when naively using two consecutive color and depth images. 
We note that misalignment is a common error in current AR applications that uses ARKit's world tracking data~\cite{dilek2018detecting}.

\subsection{Limited Sensing Capability}
Mobile vision sensors have become more potent over the past few years, especially with the newly equipped depth-sensing capability.  
However, most commercial mobile cameras still have limited FoVs ($<$ 120$^\circ$)~\cite{pixel6specs,iphone13prospecs}, much less than the desired 360$^\circ$ cameras commonly used by the movie industry to reconstruct lighting information~\cite{Li_2021_CVPR}.
Furthermore, many modern phones still do not have access to depth sensors and rely on different algorithms to estimate depth~\cite{Du2020DepthLab,zhang-indepth}.
Depth estimation errors can significantly impact the 3D point cloud generation process.
In short, the limited mobile sensing capability can make capturing high-quality data for lighting reconstruction challenging.

\subsection{Resource-quality Trade-offs}
The task of generating high-quality environment lighting information can be very compute-intensive.
State-of-the-art lighting models, which support visually coherent reflection, often require running on a powerful GPU server to achieve reasonable performance~\cite{Song2019}.
With an additional setup of physical light probes, for example, GLEAM can take from 30ms to 400ms to update scene lighting estimation~\cite{prakash2019gleam}.
By sacrificing the lighting quality, Zhao et al. achieved real-time lighting estimation (20.1ms) with low-fidelity \shc~\cite{xihe_mobisys2021};
Somanath et al. trained a deep learning model that can generate an HDR environment map in less than 9ms on recent iPhones but severely sacrifices visual coherence~\cite{Somanath2020-of}.
It is challenging to navigate the resource-quality trade-offs in providing visually coherent lighting.
 
\section{\sysname Design}
\label{sec:design}

\subsection{Overview}
\label{subsec:overview}

We design \sysname to address the challenges mentioned above to reconstruct high-quality environment lighting information. 
\sysname is an adaptive framework that progressively, e.g., as AR users naturally move around the indoor environment, reconstructs environment lighting for any user-specified reconstruction positions. 
In contrast to prior estimation-based work~\cite{Song2019,Gardner2017,pointar_eccv2020,li2020inverse}, the core of \sysname lies in how to effectively \emph{reconstruct} environment lighting information from a sequence of limited camera observations. 
Our reconstruction-based approach promises to obtain more accurate lighting information and achieves better visual results without requiring expensive data collection, model training, or physical setup~\cite{prakash2019gleam}. 

\begin{figure}
  \includegraphics[width=\linewidth]{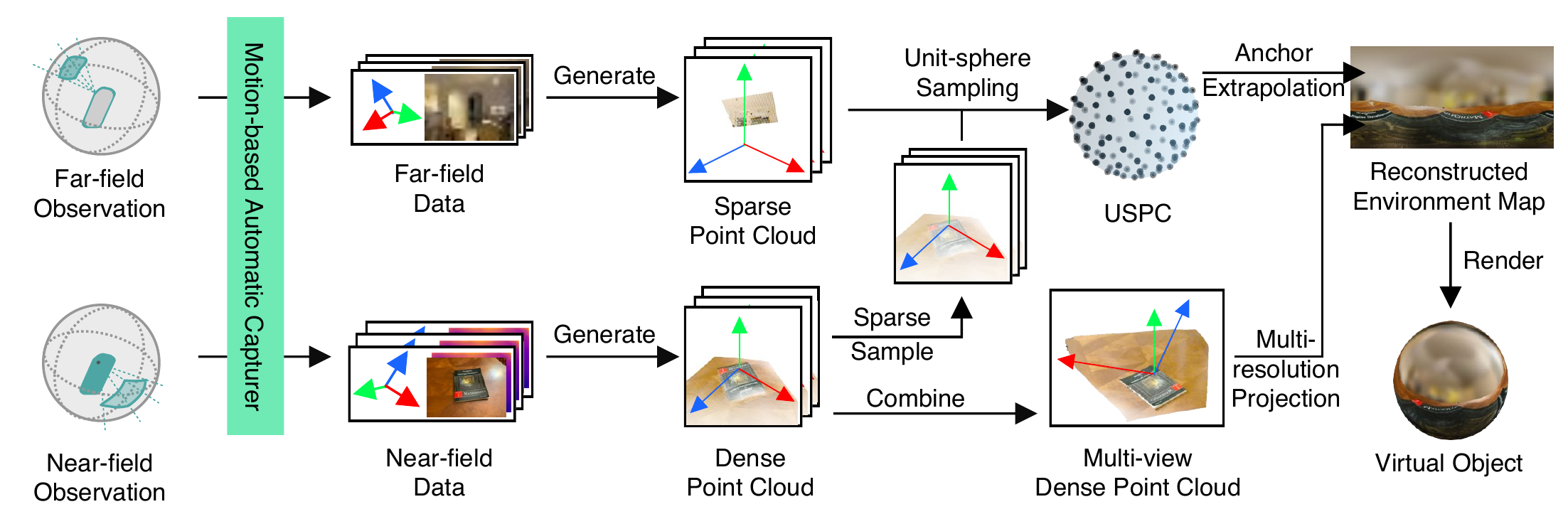}
  \caption{\sysname's overview.
    \textnormal{
        \sysname separately reconstruct for the near-field (\S\ref{subsec:spatial_variance_aware_near_field_reconstruction}) and far-field (\S\ref{subsec:distant_field_scanning}) regions from camera observations strategically guided and automatically captured. 
        The near-field data, which has a higher impact on spatial variance and consists of more accurate depth information, will be used to generate a multi-view dense point cloud. 
        The far-field data will be projected to color a unit-sphere point cloud (USPC) with the sampling technique from~\cite{xihe_mobisys2021}, and then will be extrapolated to fill neighboring pixels in the environment map.
        Finally, the dense point cloud will be projected to multi-resolution environment maps which will be combined, using our \emph{\dpcp} technique (\S\ref{subsec:dpcp}).
    }
  }
  \label{fig:reconstruction_diagram}
\end{figure}

Specifically, \sysname proposes a novel \emph{\reconstruction} technique (\S\ref{subsec:two_field_lighting_recon}) to produce geometrically accurate transformations to handle the challenge of spatial variance by transforming indirect scene observations to the desired lighting information.
Figure~\ref{fig:reconstruction_diagram} presents an overview of \sysname.
To mitigate the impact of mobility-induced noise on the reconstruction quality, \sysname proposes two policies for guiding bootstrapped device movement (\S\ref{subsec:bootstrapping_guided_startup_movement}) and automatically capturing required data based on motion (\S\ref{subsec:motion_based_automatic_capturing}).
To account for limited mobile sensing capability, \sysname only requires depth information on some camera observations (i.e., \emph{near-field observations} that have the reconstruction position in the view) and applies point cloud registration to correct the device tracking errors (\S\ref{subsec:point_cloud_registration}).
The resource intensity is dealt with from the outset with a mobile-centered design. Specifically, \sysname divides camera observations and has them go through two separate execution branches to a multi-view dense point cloud and a fixed-size point cloud.
This two-branch design effectively reduces the computational cost and memory consumption of \sysname.
We propose two novel performance optimization techniques, i.e., \dpcp (\S\ref{subsec:env_map_renderer}) and anchor extrapolation (\S\ref{subsec:anchor_extrapolation}), to render environment maps in real time at the edge.
While many knobs can impact the quality and efficiency of lighting reconstruction, \sysname allows mobile AR developers to make such trade-offs via a configurable design (\S\ref{subsubsec:session_init}).

\subsection{Two-Field Lighting Reconstruction}
\label{subsec:two_field_lighting_recon}

At the high level, our \reconstruction technique divides the task of lighting reconstruction into two sub-tasks: one that leverages depth information to produce high-quality lighting from near-field observations and one lightweight task for reconstructing lighting from far-field observations.
In other words, \sysname will generate two intermediate point clouds from camera observations for rendering environment maps.
A \emph{multi-view dense point cloud} is the outcome of judiciously applying the geometrically accurate transformation and dense sampling on near-field observations;
A \emph{unit-sphere point cloud} is the sampling outcome of the sparse point clouds from the far-field observations and the dense point clouds.
Recall that we divide the camera observations into two types, \emph{near-field observation} that includes the reconstruction position in the view and \emph{far-field observation} that does not.
As explained below, such division is based on the key insight that camera observations are subject to varying levels of spatial variance. 

\begin{figure}
    \centering
    \begin{subfigure}[b]{0.3\linewidth}
        \centering
        \includegraphics[width=\linewidth]{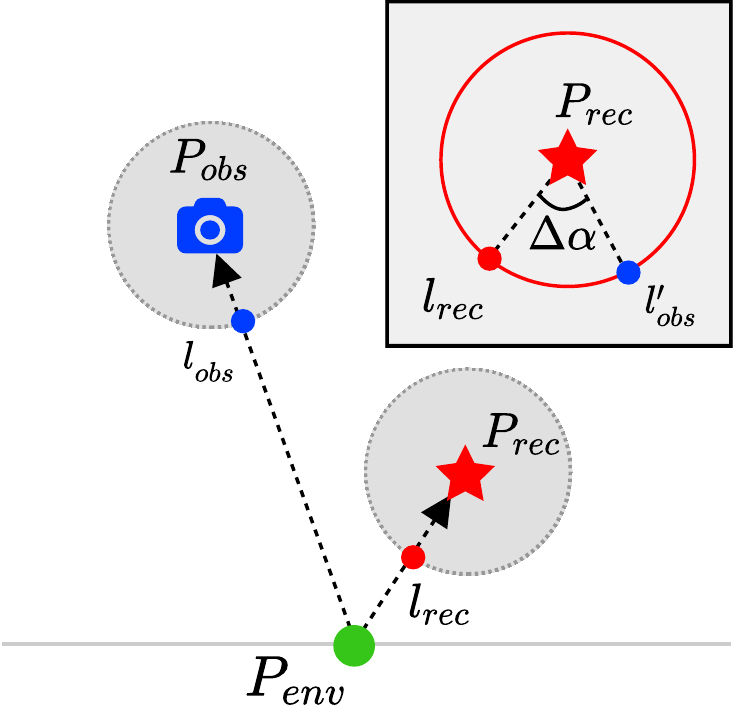}
        \caption{Spatial variance in near-field}
        \label{subfig:sv_near_fields}
    \end{subfigure}\hfill
    \begin{subfigure}[b]{0.3\linewidth}
        \centering
        \includegraphics[width=\linewidth]{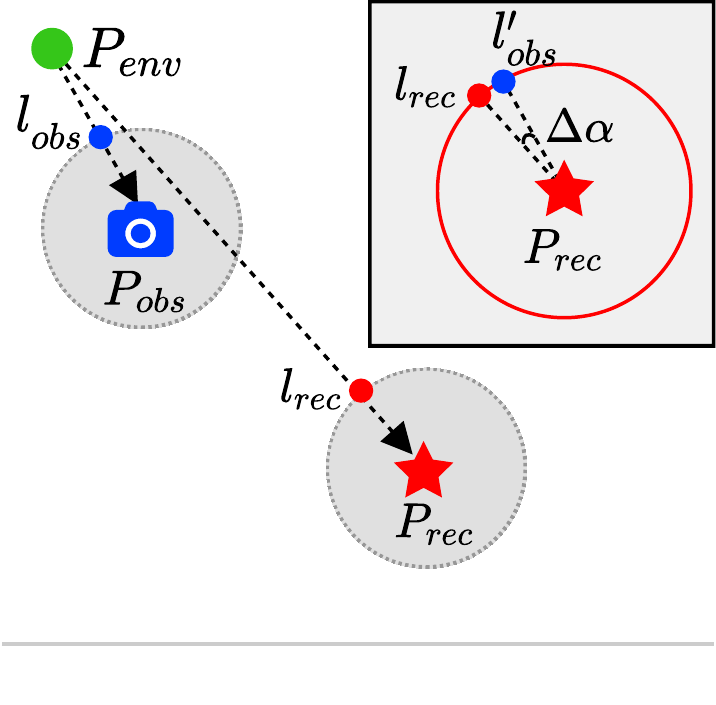}
        \caption{Spatial variance in far-field}
        \label{subfig:sv_far_fields}
    \end{subfigure}
    \hfill
    \begin{subfigure}[b]{0.3\linewidth}
        \centering
        \includegraphics[width=\linewidth]{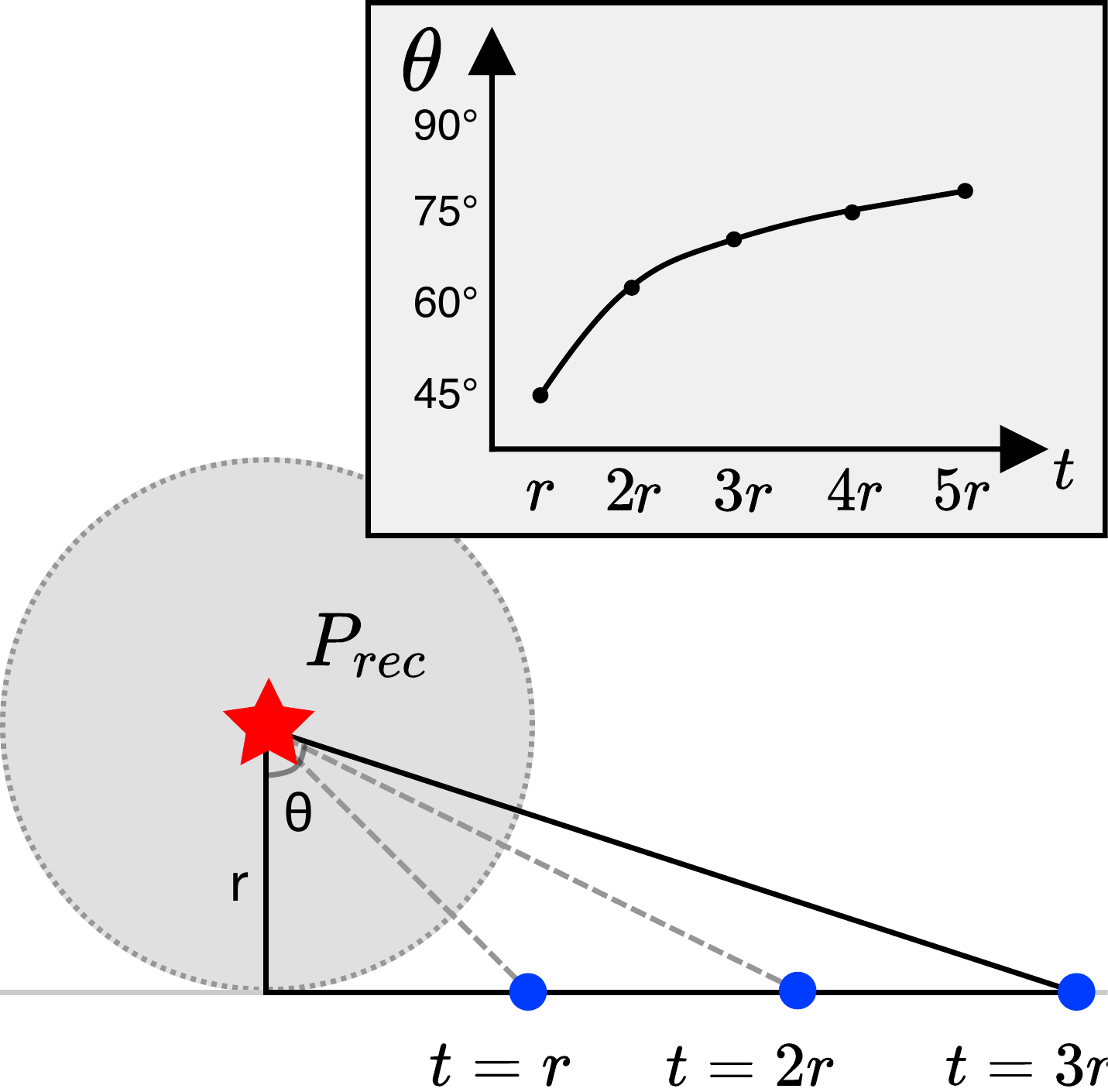}
        \caption{Near-field boundary size}
        \label{subfig:sv_near_field_projection}
    \end{subfigure}
    
  \caption{Illustration of spatial variance impact.
    \textnormal{
        We visualize the near-field and far-field observations in a 2D front view.
        $P_{obs}$ and $P_{rec}$ represent observation position and reconstruction position, respectively. 
(a) \& (b) It is more important to consider spatial variance for pixels observed in near-field than ones in far-fields, i.e., larger $\Delta\alpha$.
        (c) Increasing the near-field boundary size (e.g., from $r$ to $2r$ vs. $2r$ to $3r$) has a diminishing impact on the reflection surface.
}
  }
  \label{fig:spatial_variance_near_far_fields}
\end{figure}

Figure~\ref{fig:spatial_variance_near_far_fields} illustrates the different importance of considering spatial variance, depending on the relative position of the interested pixel to the observation and reconstruction positions.
Assume a position $P_{env}$ in the physical environment. 
To render $P_{env}$ on a virtual object surface, $P_{env}$ should be observable from the reconstruction position $P_{rec}$. 
To perceive any position $P_{env}$ in the environment, one has to observe light emitted/reflected from $P_{env}$. 
Without loss of generality, in Figure~\ref{subfig:sv_near_fields}, we show the intersection $l_{rec}$ of vector $\langle P_{env}, P_{rec} \rangle$ and the surface of the unisphere (with $P_{rec}$ being the center) represents the desired reflection. 
However, if we directly reconstruct the light ray from the camera observation position $P_{obs}$, we end up with $l_{obs}$, the intersection between vector $\langle P_{env}, P_{obs} \rangle$ and the surface of the $P_{obs}$-centered unisphere. 
If we translate $l_{obs}$ to the $P_{rec}$-centered unisphere by applying the vector $\langle P_{obs}, P_{rec} \rangle$, we will get a third intersection $l'_{obs}$. 
Observe that the light ray represented by $\langle P_{rec}, l'_{obs} \rangle$ can differ significantly (i.e., larger $\Delta\alpha$) from $\langle P_{rec}, l_{rec} \rangle$, the light ray that should be perceived at $P_{rec}$.
In short, near-field observations are impacted much more by spatial variance.
On the contrary, as shown in Figure~\ref{subfig:sv_far_fields}, if the $P_{env}$ is in the far-field, $l'_{obs}$ can be a good approximation for $l_{rec}$ (i.e., much smaller $\Delta\alpha$).
In short, far-field observations are impacted much less by spatial variance.

Other benefits of separating camera observations into near-field and far-field include better tolerance of limited mobile depth-sensing capability and resource efficiency.
A naive alternative design of applying the same reconstruction pipeline to all camera observations can lead to incorrect point clouds and demand resources proportional to the indoor scene space (i.e., the total number of points). 
In contrast, our design of processing far-field observations demands less memory and computation resources by using a fixed-size point cloud~\cite{xihe_mobisys2021}.

\subsubsection{Spatial Variance-Aware Near-Field Reconstruction}
\label{subsec:spatial_variance_aware_near_field_reconstruction}

To effectively transform camera observation(s) to the environment map at the reconstruction position, \sysname leverages the increasingly popular depth sensor in mobile camera system~\cite{Huawei_undated-ca,ipad-lidar}.
Depth sensors enable the possibility of capturing geometrically accurate environment observations. 
First, we densely sample color and depth buffers of the camera images for each near-field observation.
Then, the image buffers and the camera transformation matrix (i.e., rotation and translation) are used to generate the geometry and color of a dense 3D point cloud.
In our implementation, we use the camera transformation matrix information provided by commercial AR frameworks. Such information is often referred to as device tracking data. 
The point cloud generation process can be time-consuming. To speed up this process, we separate the geometry and color generation tasks and then execute both tasks on the GPU.
Finally, the output dense point cloud (i.e., geometry and color information) is written to a global point cloud buffer that maintains a multi-view point cloud for the current reconstruction session.

To support multi-view reconstruction, \sysname uses a motion-based automatic capturing scheme (\S\ref{subsec:motion_based_automatic_capturing}) to supply the reconstruction pipeline with new near-field observation. 
\sysname assigns a unique indexing identifier for each near-field observation. 
This identifier is subsequently used for other data, including the camera transformation matrix and the derived dense point cloud.
The global point cloud buffer is updated with the least-recently observed policy, i.e., the points associated with the oldest near-field observation will be replaced.

We define \emph{near-field boundary} as a constrained cubic space that contains points belonging to the near-field observations that will undergo further processing.
For points outside the near-field boundary, we will not perform \dpcp (\S\ref{subsec:dpcp}).
Theoretically, this boundary can be as big as the indoor scene. 
However, having a too large boundary may have undesirable implications on both the memory and computation consumption and is often unnecessary.
Figure~\ref{subfig:sv_near_field_projection} shows that increasing the near-field boundary size has a diminishing return on the reflection surface. 
For example, increasing the boundary from 2X to 3X of the virtual object size only increases the coverage by 8.13$^\circ$, a 1.13X. 
In short, we use a configurable near-field boundary for a trade-off between resource consumption and reconstruction quality.
We set the near-field boundary side length to two meters based on the AR virtual objects we use for testing. 
It is worth noting that AR developers should adjust this boundary for large virtual objects or divide the large object into smaller objects to have multiple reconstruction positions. 
The dense near-field point cloud generation allows us to produce geometrically accurate camera observation transformations and produce continuous lighting representations.

Our spatial variance-aware observation transformation still generates an approximated observation at the reconstruction position due to the lighting variance from different observation directions.
We assume that the light does not change between different observation directions, which suffices for mobile AR rendering in most cases.
However, such an assumption may lead to visually incorrect results if the environment around the reconstruction position contains reflective physical objects, e.g., mirrors.
It is possible to further address such concerns by meticulously selecting light ray directions between observations, which we leave as future work.

\subsubsection{Directional-Aware Far-Field Reconstruction}
\label{subsec:distant_field_scanning}

Mobile depth sensors usually only capture the surrounding environment within a limited range. For example, the LiDAR sensor on iPhone 13 Pro can capture depth up to 5 meters away~\cite{iphonelidar_range}. Thus, it might not be suitable for sensing large physical environments.  
However, the desired environment lighting for AR rendering varies directionally depending on the surrounding physical world environment.
To address the directional variations, i.e., the anisotropy of environment lighting, LitAR reconstructs far-field lighting by sparsely sampling camera observation to provide omnidirectional lighting.

Even with the recent advancements in hardware, modern cameras still have small FoVs and thus capture only a small portion of the environment covering limited directions. 
However, reconstructing a dense point cloud to address the anisotropic lighting property from far-field observations is impractical as generating a dense point cloud for a far-field environment can be potentially unconstrained regarding computation and data storage.
In addition, objects in the far-field observations may exceed the range limit of mobile depth sensors, which makes it difficult to obtain geometrically accurate transformation. 
The inherent nature of the far-field environment thus leads to lower confident far-field depth observations, which makes it ill-suited for dense point cloud reconstruction.

To address these limitations, we design a lightweight process to reconstruct far-field lighting from sparsely sampled camera images. 
Recall that a camera observation is considered a far-field observation if the reconstruction position falls outside the camera view. 
For a far-field observation, we sparsely sample a low-resolution camera image and obtain the current camera transformation matrix, similar to \S\ref{subsec:spatial_variance_aware_near_field_reconstruction}.
Note that we do not capture depth data for the far-field observation as its depth information may be inaccurate, and the spatial variance has less impact.

To generate the sparse point cloud, we assume the depth of all pixels to be one and scale the camera intrinsic values accordingly.
We use a similar design to \cite{xihe_mobisys2021} by projecting the sparse point cloud to a set of uniformly distributed points, referred to \emph{anchors}, on a unit sphere.
The resulting data structure is a \emph{\uspc} (USPC). 
As demonstrated in prior work~\cite{xihe_mobisys2021}, the design of USPC is aware of directional lighting variance and thus addresses the anisotropy property of environmental lighting.
In this work, we set the number of anchors of the USPC to be $1280$, the same as prior work~\cite{xihe_mobisys2021}.
The anchor points are colored by combining the color data from the sparse point cloud and ambient light sensor readings. 
Recall that we want USPC to represent the lighting from all directions, including near-field observations. 
Therefore, we sparsely sample the dense point cloud generated from the near-field observation; then, the resulting sparse point cloud is similarly projected to the same USPC.
In short, the reconstructed far-field lighting is represented as a unit-sphere point cloud with a much smaller memory footprint (proportional to the anchor size) while still providing sufficient directional-aware lighting information.

\subsection{Noise-tolerant Data Capturing Policies}

\subsubsection{Guided Bootstrapped Movement}
\label{subsec:bootstrapping_guided_startup_movement}

Another key design of \sysname to reconstruct high-quality lighting is to exploit user movement, a feature of mobile AR.
We observe that commercial mobile AR frameworks such as ARKit have built-in support for explicitly guiding mobile AR users to scan their physical surrounding environment before using the app. 
Note such practice is often used for calibrating world tracking data, but not for lighting estimation~\cite{arkitcoachingoverlay}. 
However, this commonly adopted movement practice typically leads to a biased sampling of environment lighting in concentrated observation directions.
This biased sampling is due to the narrow focus on increasing the observation of the nearby environment around the reconstruction position. 
Although commercial frameworks use deep learning-based models to estimate environment lighting from observations, such biased observations create a barrier to more accurate estimation.
As we will show in \S\ref{subsec:eval_near_field_recon}, increasing observations with the common practice shows little improvement in rendering results.

\begin{figure}
\centering
    \begin{subfigure}[b]{0.29\linewidth}
        \centering
        \includegraphics[width=0.85\linewidth]{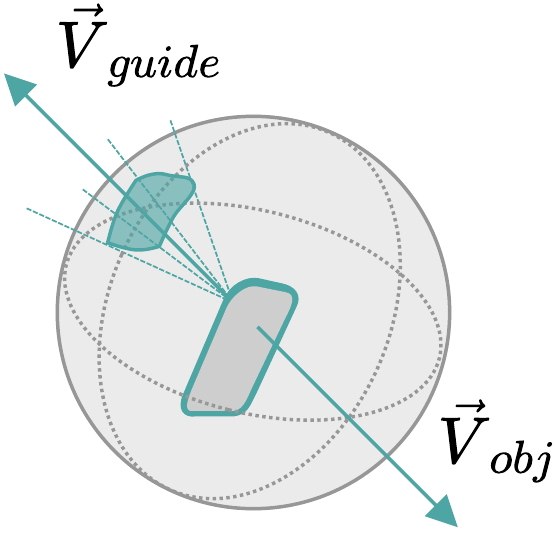}
        \caption{Front \& back facing observations}
        \label{subfig:front_and_back_facing_obs}
    \end{subfigure}\quad
    \begin{subfigure}[b]{0.3\linewidth}
        \centering
        \includegraphics[width=0.75\linewidth]{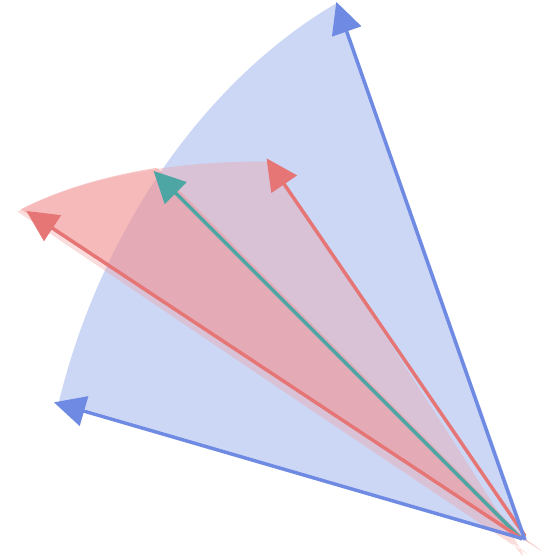}
        \caption{Horizontal \& vertical movement}
        \label{subfig:horizontal_and_vertical_mov}
    \end{subfigure}
    \quad
    \begin{subfigure}[b]{0.36\linewidth}
        \centering
        \includegraphics[width=\linewidth]{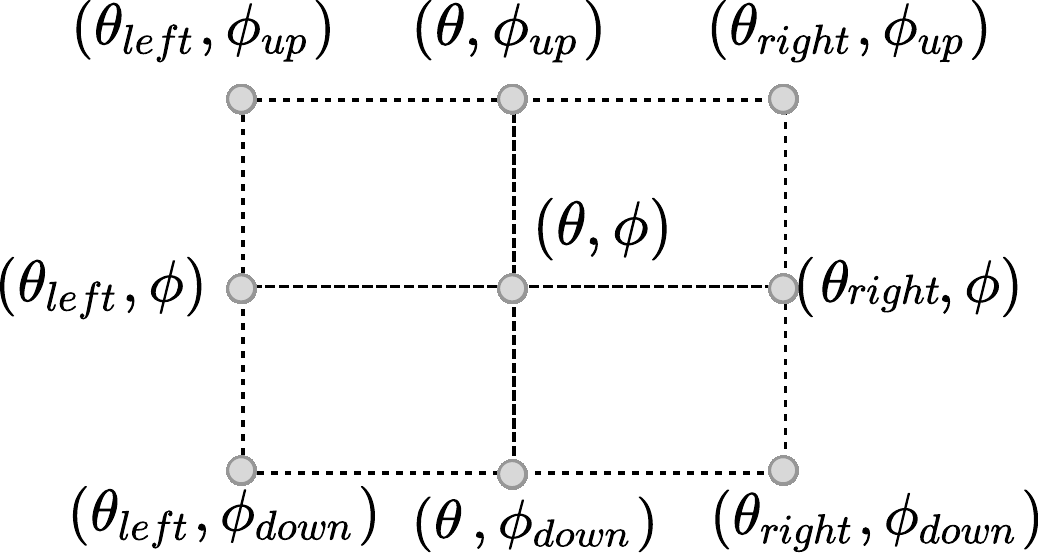}
        \caption{Movement direction grid}
        \label{subfig:movement_grid}
    \end{subfigure}
    \caption{Guided bootstrapped movement.
        \textnormal{
            Our guided bootstrapping movement technique directs users to point the mobile camera toward the opposite direction $\vec{V}_{guide}$ of the virtual object viewing direction $\vec{V}_{obj}$ and rotate along horizontal and vertical directions to increase the far-field observations. 
            The movement follows a grid-style pattern, where ($\theta$ and $\phi$) denote the observation direction in spherical coordinates.
        }
    }
  \label{fig:guided_movement_diagram}
\end{figure}

Instead, we propose a novel yet simple guided movement policy to look at the \emph{backward environment}, i.e., observable from the opposite direction to the virtual object viewing direction. 
This guided movement is designed to increase the observation directions rather than the observation overlapping and to help address the anisotropic lighting property.
In other words, our guided movement policy provides bootstrapped data at the AR application startup time to increase the far-field observations.
As illustrated in Figure~\ref{fig:guided_movement_diagram}, our policy guides the user to look at the backside of the intended virtual object observation direction.
We use a grid-style pattern to partition the observation directions and provide guided viewing directions for the user.
Specifically, as shown in Figure~\ref{subfig:movement_grid}, the optimal choices on the number of observations could be $\{1,3,5,9\}$ based on the horizontal and vertical direction partitions.
The number of observations corresponds to the combination of moving \emph{left} and \emph{right} in horizontal direction, as well as \emph{up} and \emph{down} in vertical direction.
We currently use an angular difference of 30 degrees in horizontal and vertical directions.
Furthermore, as our far-field reconstruction method does not reconstruct detailed observations, our guided movement can be performed without the user focusing on each camera observation orientation for a long time.
In \S~\ref{subsec:eval_far_field}, we will show that by using the guided movement, \sysname can find more accurate color tones to fill unseen areas and produce more accurate renderings.

\subsubsection{Motion-based Automatic Capturing.}
\label{subsec:motion_based_automatic_capturing}

Continuously capturing all camera observations or relying on the mobile AR user to manually capture them can lead to poor usability, low-quality data (e.g., images with motion blur), and high consumption of mobile resources. 
For example, prior work has demonstrated that motion blur is a common occurrence in mobile AR---which we also observe---and can lead to low accuracy for AR tasks~\cite{Liu2020-fy}.
Additionally, a recent low-frequency lighting estimation framework has demonstrated that strategically skipping the capture of specific camera frames has little impact on the estimation accuracy~\cite{xihe_mobisys2021}.
Intending to provide good usability, capture high-quality data, and reduce resource consumption, we design a \emph{motion-based automatic capturing} technique that leverages multi-sensor data to automatically select camera frames and AR data for lighting reconstruction.
In a nutshell, this technique will only capture observation data that is new spatially (i.e., by checking device position and rotation information) and temporally (i.e., by updating a previously captured frame with the same device information).

Specifically, \sysname uses a simple timer-based policy to assess the need to capture new data by checking if the mobile device has exhibited significant movement. 
In this work, we leverage the device accelerometer and gyroscope sensors to maintain a moving window of the device's most recent $K$ position and rotation information (i.e., 6DoF). 
Every $C$ milliseconds, \sysname will compare the device 6DoF information at the current frame to the ones in the moving window to assess the likelihood of motion blur. 
If the device pose has changed for more than 10cm and 10$^\circ$, the device is considered to have significant movement in a short time window, and the current frame is skipped.
\sysname will re-run the check every frame until a new frame is found while the device is relatively stable. 
Otherwise, the current frame is captured. 
The timer will be reset once a new frame is captured in both cases. 
Our motion-based automatic capturing will produce new data at least every $C$ millisecond, depending on the mobility. 
Both the moving window and capturing frequency can be configured. In our implementation, we set $K$ = 5 and $C$ = 300.

\subsection{Real-time Environment Map Rendering Techniques}
\label{subsec:env_map_renderer}

Thus far, our \reconstruction has generated two intermediate point clouds.
To support high-quality multi-view lighting reconstruction for mobile AR, we need to convert the point cloud presentation into a lighting representation commonly supported by modern rendering engines. 
In this work, we choose \emph{environment map} as the final lighting representation, which most mobile rendering engines can directly use.
At the high level, to generate a high-quality environment map (i.e.,  visually continuous pixels) from a point cloud that consists of discrete points, one often needs to handle occlusion and inter-point connection.
One common way to recreate the inter-point connections and calculate occlusion is to resort to conventional 3D reconstruction methods, e.g., Ball-Pivoting surface mesh reconstruction algorithms~\cite{bernardini-ball-pivoting}.
However, such a method is ill-suited for real-time applications as surface mesh reconstruction can be computationally expensive, e.g., we observe that it takes about 3 seconds to perform mesh reconstruction on a five-view dense point cloud.
In this section, we describe two novel techniques called \emph{\dpcp} and \emph{\extrapolate} for generating near-field and far-field portions of the environment map in real-time.

\begin{figure}
\centering
    \begin{subfigure}[b]{0.78\linewidth}
        \centering
        \includegraphics[width=\linewidth]{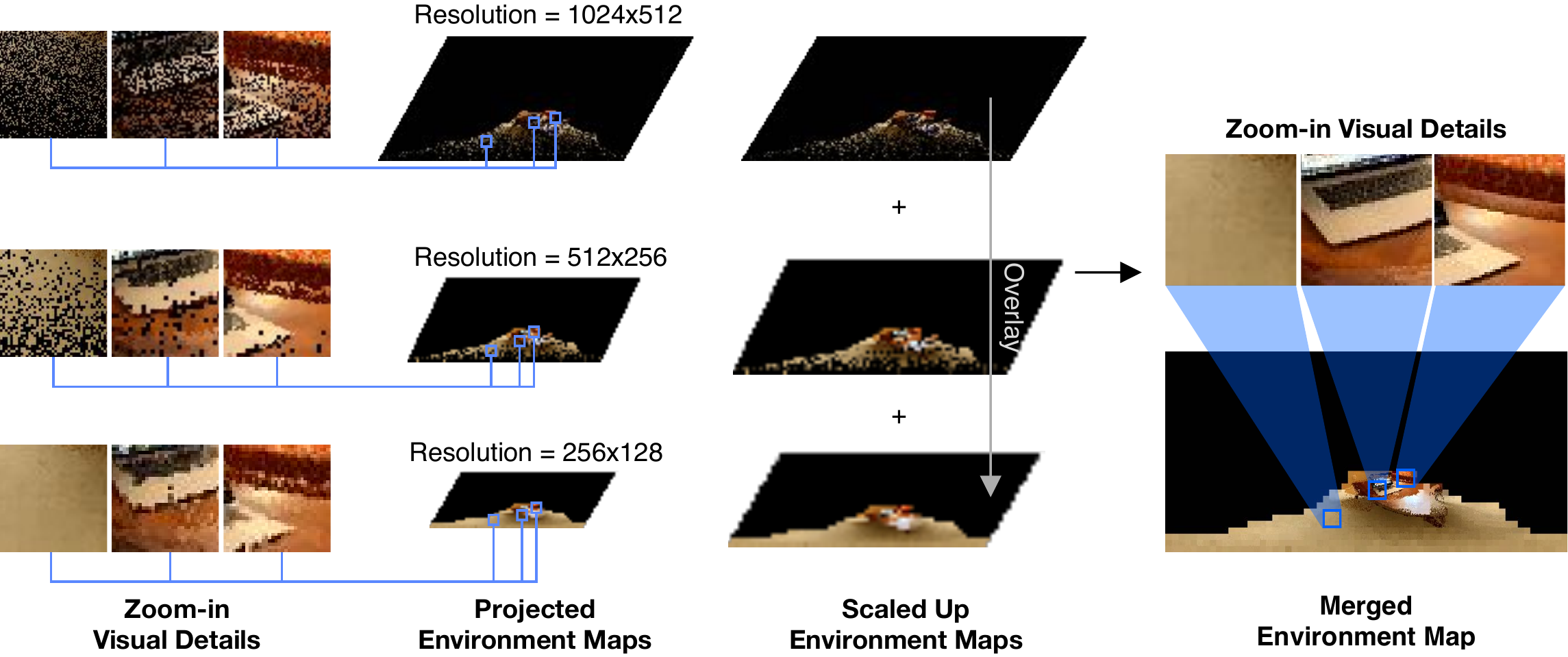}
        \caption{Multi-resolution projection}
        \label{subfig:illustration_dpcp}
    \end{subfigure}\hfill
    \begin{subfigure}[b]{0.2\linewidth}
        \centering
        \includegraphics[width=\linewidth]{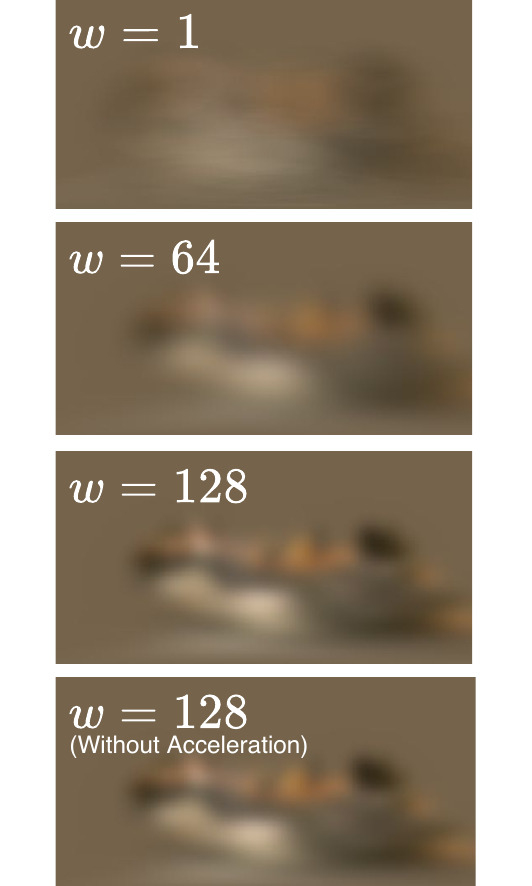}
        \caption{Anchor extrapolation}
        \label{subfig:illustration_ae}
    \end{subfigure}
    
    \caption{Illustration of \dpcp and anchor extrapolation.
        \textnormal{
        (a) We show an example of \dpcp with three levels. 
        The dense point cloud will be projected three times.
        The higher projection resolution leads to more visual details but has more discrete pixels; the lower projection resolution has more continuous pixels but fewer visual details and exhibits pixelation.
        We obtain a high-quality final environment map by scaling and overlaying each intermediate environment map.
        (b) We show that a larger $w$ value leads to a less blurry environment map and that our nearest-anchor acceleration has minimal visual impact. 
        }
    }
  \label{fig:env_map_rendering}
\end{figure}

\subsubsection{Multi-resolution Projection.}
\label{subsec:dpcp}

We propose a lightweight technique called \dpcp to convert the near-field dense point cloud to the respective portion in the environment map.
We use the common equirectangular format to present the environment map.
Specifically, \dpcp projects a point cloud into a set of environment map images with decreasing resolutions and addresses the inter-point connection and occlusion at the 2D-pixel level.
When projecting a point cloud onto one environment map image, \dpcp first converts the position of the point cloud from the Cartesian coordinate system to the Spherical coordinate system.
Then, for each point, we calculate its 2D projection coordinate on the environment map based on the angle values of its spherical coordinates.
Then, we assign the point cloud color to the corresponding pixel on the environment map.
As multiple points can be projected to the same pixel of the environment map, for each pixel, we handle the point occlusion by selecting the shortest-distant projected point to color the pixel.
Figure~\ref{subfig:illustration_dpcp} illustrates an example of three-level projection.

However, when the point cloud density is low, e.g., due to low capturing resolution, projecting point cloud only onto one environment map image resolution may lead to degraded visual quality. 
For example, it might result in an image with discretely projected points rather than a continuous view of the scene, and it might not adequately represent the inter-point occlusion.
To address these issues, we assign different size values for projected points via multi-resolution image projection.
We first project the point cloud into a series of images with decreasing resolutions.
Then we scale all the projected images to the largest resolution via the nearest pixel interpolation.
Finally, the \dpcp results are merged into a single environment map by selecting the shortest-distant projected point to the reconstruction position from each projected image per pixel.
If multiple projections have the same distance, we select the one from the highest resolution as it has more visual details.

We note that the number of resolution levels and per-level resolution can be adjusted for different combinations of dense point clouds and reconstruction positions. 
However, our design of the near-field boundary described previously in \S\ref{subsec:spatial_variance_aware_near_field_reconstruction} suggests that all reconstructed near-field dense point clouds will be confined to a cubic space. 
Thus, it is possible to have a relatively fixed configuration to handle various scenes.
In this work, we choose two resolution levels with per-level resolution as 1024x512 and 512x256, unless otherwise specified.

\subsubsection{Anchor Extrapolation.}
\label{subsec:anchor_extrapolation}
Recall that by now, our \reconstruction has generated a colored \uspc (USPC) for the far-field lighting.
To generate the corresponding environment map in the equirectangular format, we use the anchor points to color each environment map pixel. 
However, the USPC, by design, only has a fixed number of anchor points.
Therefore, directly projecting anchor points to the environment map is likely to lead to many empty pixel values. 
To address this problem, we design an \emph{anchor extrapolation} technique that calculates each pixel value as a weighted average of USPC anchor values. 
This technique, in essence, assigns color value to pixels by extrapolating from their nearby anchor colors and will result in a gradient coloring and blurring effect. 

Specifically, we first initialize each pixel of the environment map with a normal vector, i.e., a unit vector from the sphere center to the pixel position. 
The initialization is feasible as a pixel in the equirectangular format of an environment map can be easily presented in the spherical coordinate system.
We then calculate the $i^{th}$ pixel color $c_i$ using the following equation: 
\begin{equation}
    c_i =\frac{2}{N}\sum^{N}_{j=1}{\max(\vec{p}_j \cdot \vec{n}_i, 0)^w c_j},
    \label{eq:anchor}
\end{equation}
where $\vec{n}_i$ represents the pixel normal vector, $N$ is the number of anchors, $\vec{p}_j$ and $c_j$ are the normal vector and color for the $j$-th anchor, respectively.
Note that the dot product between $\vec{p}_j \cdot \vec{n}_i$ is effectively the cosine value of the angle between these vectors, as $|\vec{p}_j|=|\vec{n}_i| = 1$.
The $\max$ function effectively filters out all the anchor points in the hemisphere opposite the $i^{th}$ pixel. Furthermore, $w$ is an exponent controlling the blurring level of far-field reconstruction.
Intuitively, a smaller $w$ value will lead to more anchor points used for the pixel calculation.
Thus, a smaller $w$ value will result in a blurrier environment map, while a larger $w$ will produce a clearer environment map, as demonstrated in Figure~\ref{subfig:illustration_ae}.
In this work, we set $w$ to be 128.

Note that calculating the pixel color using Equation~\eqref{eq:anchor} can be time-consuming as the weighted average has to iterate through all anchor points. 
However, anchors do not contribute equally to the pixel color calculation. 
Intuitively, an anchor $j$ that has a smaller $\max(\vec{p}_j \cdot \vec{n}_i, 0)$ decreases more quickly with the power $w$.
Such anchors are also farther away from the pixel of interest than anchors with a larger $\max(\vec{p}_j \cdot \vec{n}_i, 0)$ value. 
In fact, we find that when $w=128$, only the 32 nearest anchors out of the 1280 contribute significantly (i.e., $\max(\vec{p}_j \cdot \vec{n}_i, 0)$ > 0.1). 
Thus, to speed up the pixel color calculation, we precompute the 32 nearest anchors for each pixel and their respective cosine values.
The precomputation effectively reduces the number of anchors by a factor of 40 and allows the use of cached results for the weighted average calculation. 
Figure~\ref{subfig:illustration_ae} shows that our acceleration has minimal visual impact.

\subsection{\sysname Quality-Performance Configurations}

\subsubsection{Reconstruction Session Settings and Initialization}
\label{subsubsec:session_init}

\sysname uses a \emph{\lrs} to manage each multi-view reconstruction task.
A \lrs has the same lifecycle as its corresponding virtual object; 
the session is created when a virtual object placement request is issued and is destroyed when the placed object is removed from the scene.
As AR applications might need multiple virtual objects in the view, \sysname supports multiple active \lrs per AR session (i.e., during the AR application's lifetime).
At the beginning of each session, \sysname collects static device-specific information, e.g., camera intrinsic, current ambient lighting data, and camera image native resolutions, to bootstrap subsequent lighting reconstruction operations.

\sysname supports configuring several knobs, including color image sampling rate, number of views, \dpcp resolution levels, and environment map size, that trade-off visual quality and reconstruction performance. 
These knobs can be categorized into three types, i.e., data capturing, \reconstruction, and environment map rendering.
Thus, the startup latency of each session and the subsequent near/far-field reconstruction depend on the specific configurations. 
The users (e.g., mobile AR developers) can configure each \lrs based on performance requirements or select one of the three presets: low, medium, and high.
In \S\ref{subsec:eval_quality_perf_trade_offs}, we will show that all three presets achieve better visual quality than ARKit but take an increasing amount of time to generate an environment map.

\subsubsection{Point Cloud Management.}
\label{subsec:point_cloud_registration}

To achieve low-latency point cloud operations, \sysname leverages the edge to generate, manage, and transform both the sparse and multi-view dense point clouds. 
To exploit the inherent parallelism of point cloud operations, \sysname performs these operations on the GPU. 
However, even with unified memory, the managed memory still must be copied to the GPU memory (by the driver) for data access. 
A naive implementation may lead to expensive GPU memory access overhead. 
Thus, we carefully design the memory layout using a continuous memory buffer to store the multi-view dense point cloud and a fixed number of anchor points.
When new view data is processed, \sysname overwrites the point cloud memory buffer by replacing the oldest data for \emph{temporal} consistency or replacing the data with the same view identifier for \emph{spatial} consistency.
This fixed-view design keeps the memory layout unchanged, thus avoiding paging setup overhead while still producing high-quality environment maps.

Additionally, \sysname includes an asynchronous point cloud registration to address the mobility-induced noises, which can lead to misaligned point positions. 
In other words, \sysname runs point cloud registration in parallel to the main \reconstruction and will update the environment map with the aligned point cloud once the registration completes.
We note that the point misalignment is mainly due to the inaccurate device tracking data provided by the AR framework, in this case, ARKit.
Providing accurate device tracking information is an essential but orthogonal research question; 
prior work such as ORB-SLAM2~\cite{murORB2} and Edge-SLAM~\cite{Ben_Ali2020-ri} can achieve good tracking in about 26ms-50ms.
In this work, we use the iterative closest point registration~\cite{besl1992method} to mitigate the impact of noisy tracking data on the lighting reconstruction. 
During our preliminary study, we found that point cloud registration is not always necessary (e.g., when mobile AR users are relatively static) and can take significantly longer than other operations (e.g., 200ms for handling five views with 1024x768 points).
In our implementation, the point cloud registration is turned off by default.
 
\section{Implementations}
\label{sec:implementations}

\begin{figure}
\centering
\includegraphics[width=0.9\linewidth]{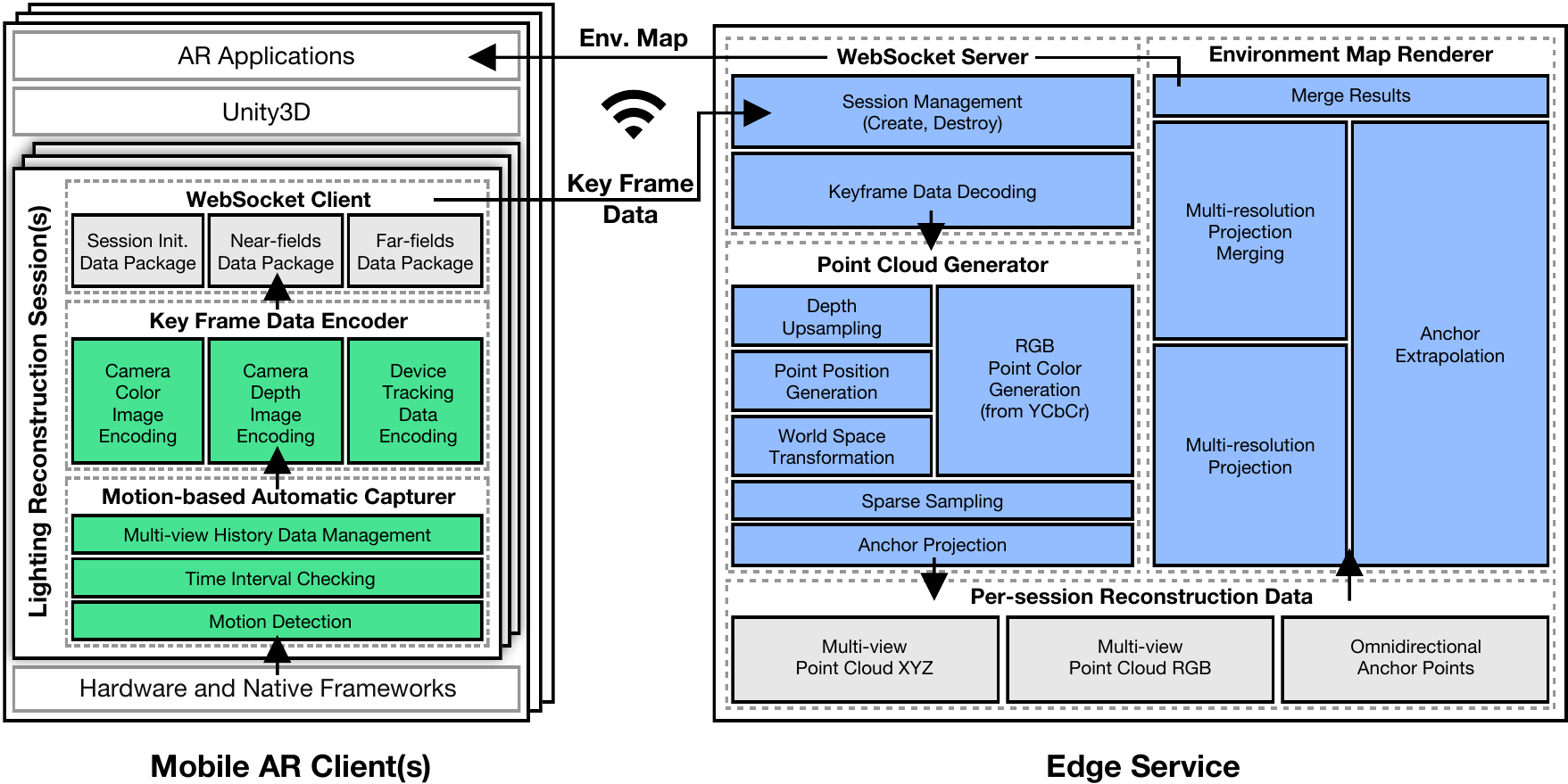}
\caption{A system architecture overview of \sysname.
    \textnormal{
        The framework provides high-quality lighting as an environment map for mobile AR applications with two logical components: client-side and edge service. 
        \sysname can support lighting for multiple reconstruction positions per AR session on the client-side and allows multi-user scene data sharing via the edge service. 
    }
  }
  \label{fig:system_architecture}
\end{figure}

We implement all the techniques described in \S\ref{sec:design} in a prototype system called \sysname (\S\ref{subsec:system_implementation}). 
To facilitate experiments in a controlled manner, we also develop a mobile AR simulator based on Unreal Engine (\S\ref{subsec:simulator}).

\subsection{Edge-Based Prototype}
\label{subsec:system_implementation}

We implement \sysname in C\# and Python with about $2.2$K lines of code. 
Figure~\ref{fig:system_architecture} shows an architecture overview of \sysname.
\sysname consists of two logical modules: a client-side that captures, encodes, and sends the data necessary for the lighting reconstruction sessions; an edge-side that decodes the keyframe data, generates intermediate point clouds, and renders the environment map. 
\sysname currently supports AR applications built with Unity3D~\cite{unity3d} and AR Foundation~\cite{arfoundation}.

\subsubsection{Client-side.}
The client module of \sysname is implemented as a Unity package. 
We provide an entry script called \emph{ARLightingReconstructionManager} as a \emph{MonoBehavior} subclass to allow the developer to specify system configurations and visual quality-related information via the Unity editor UI.
\emph{ARLightingReconstructionManager} also manages the memory usage and function calls of all lighting reconstruction sessions in an application's lifecycle.
We implement the \emph{motion-based automatic capturer} by leveraging the AR device tracking data provided by AR Foundation to automatically capture environment data that will not be subjected to camera motion blurs. 
We perform the nearest neighbor sampling on the device's native color and depth image data for the visual data.
Specifically, for the AR application running on our testing device, iPad Pro, we sample color and depth images in the format of YCbCr 4:2:0 and float32, respectively.
The captured color and depth images are then encoded with a one-byte unique package identifier and device tracking information into three binary data packages, i.e., session initialization, near-field, and far-field. 
We refer to these data packages as \emph{keyframe data}.
\sysname also manages a WebSocket session for low latency communication with the edge server, i.e., sending keyframe data and receiving environment map.

\subsubsection{Edge-side.}
On the edge side, we implement a Tornado-based \cite{tornadoframework} WebSocket service to communicate with the client.
The WebSocket server dispatches the package to different operations for each received binary data package based on its package identifier.
We leverage NumPy~\cite{harris2020array} to decode and convert the received binary packages into different data types and structures.
To improve the performance of point cloud-related operations, including point cloud generation, \dpcp, and anchor extrapolation, we implement them in CUDA kernels using the Numba library~\cite{lam-numba}.
As such, these operations run on the edge GPU. 
In our edge system memory management, we leverage the unified memory~\cite{cudaunifiedmemeory} to avoid time-consuming data copying between CPU and GPU.
Our edge server implementation can also fall back to traditional GPU memory without any modification to support other GPU hardware that does not have unified memory.

\subsubsection{Mobile-edge communication.}
We design a low-overhead and compact networking communication scheme to support the goal of low-latency lighting reconstruction.
Both the near and far-field data are serialized into binary formats. 
Specifically, we stream only the device's native color and depth data for near-field data and reconstruct the point cloud on the edge side.
Compared to directly streaming float32-encoded point cloud in the XYZRGB format, we need at most 22.9\% of the bytes.
Also, as we only capture sparse camera images during far-field reconstruction, the far-field keyframe data is significantly smaller than the near-field counterpart (1189 bytes vs. 270389 bytes).

\begin{figure}[t]
\includegraphics[width=\linewidth]{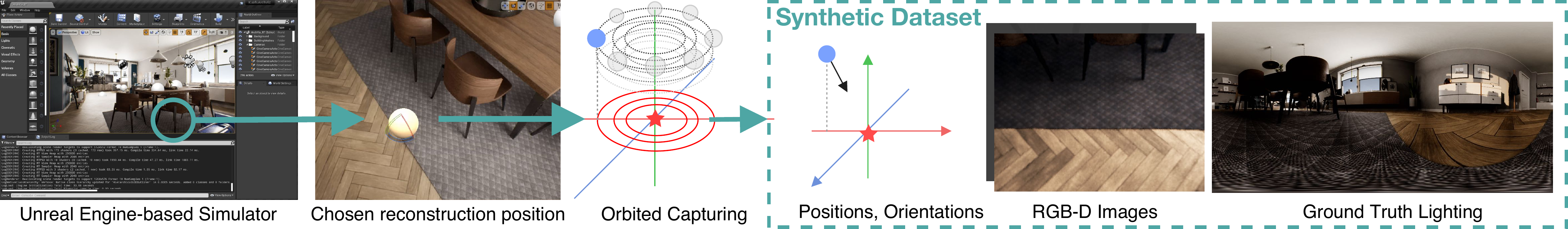}
\caption{Unreal engine-based simulator and synthetic dataset generation process.
    \textnormal{
        We generate an indoor synthetic dataset using an Unreal Engine-based simulator.
        We use an orbit trajectory at each manually chosen position to build observations around the position and place the virtual camera to capture images. 
        We extract camera positions, orientations, RGB-D images, and ground truth environment map.
    }
  }
  \label{fig:simulation_exp_workflow}
\end{figure}

\subsection{Unreal Engine-Based Mobile AR Simulator}
\label{subsec:simulator}

We implement a mobile AR simulator by leveraging a high-fidelity 3D graphics rendering engine, Unreal Engine~\cite{ue4}. 
Figure~\ref{fig:simulation_exp_workflow} presents the workflow of our simulator.
First, we use Unreal Engine 4 to create a photorealistic indoor scene based on the \emph{ArchViz}\footnote{https://docs.unrealengine.com/4.27/en-US/Resources/Showcases/ArchVisInterior/} project, which provides high-quality architectural visualization for interior design.
Next, we create a virtual camera using the Blueprints Visual Scripting\footnote{https://docs.unrealengine.com/5.0/en-US/blueprints-visual-scripting-in-unreal-engine/} system that takes controlled variables to modify the camera's movement and internal properties, e.g., FoV.
We simulate the device/user movement by moving the virtual camera within a photorealistic 3D indoor scene. 
Finally, we can configure the simulator with desired variable values to generate a synthetic dataset for rendering purposes, including camera observations and environment physical properties.
Our simulator can be extended to support future studies and bears the following advantages over a device-based setup:
\1 our simulator makes it easier to extract high-quality environment lighting and physical information to serve as the ground truth. At the same time, obtaining such information can be expensive or unpractical due to physical limitations on measurement and observation.
\2 it is easier to study individual factors in isolation by applying controlled changes to the scene environment and simulated mobile devices.

\subsubsection{AR Virtual Object Rendering.}
We develop a browser-based renderer using the Three.js rendering framework~\cite{threejs} to automate the process of rendering virtual objects of interest. 
Specifically, our renderer uses information, including reconstructed lighting, the camera position, and properties, from our synthetic dataset to render a 3D virtual object at the resolution of 1024x768. 
The renderer then trims empty pixels outside rendered objects to remove the object-to-frame size impact on PSNR calculation when using different camera FoV settings.
The resulting images of rendered objects serve as the basis for comparing different lighting reconstruction methods.
 
\section{Evaluation}
\label{sec:eval}

We evaluate the performance of \sysname using a lab testbed and the simulator.
The lab testbed includes a LiDAR-enabled iPad Pro serving as the client and a Jetson Xavier NX~\cite{jetsonagxxavier} board serving as the edge server.
The iPad and the Jetson board communicate via resident WiFi with an average latency of 7.08 ms ($\pm$ 3.31 ms)  and network bandwidth of 508 Mbits/sec ($\pm$ 12 Mbits/sec).
For testbed-based experiments, we choose three different indoor scenes and compare \sysname with three different baselines:
\1 ARKit 5~\cite{arkit}, a commercial AR framework developed by Apple; 
\2 \sysname with \emph{point cloud registration} turned on; 
\3 \sysname with \emph{mesh reconstruction} instead of the lightweight \dpcp module. 
We use the Environment Probe~\cite{arkit-envprobe} feature of ARKit to generate environment maps.
The lighting estimation feature of ARKit is backed up by EnvMapNet~\cite{Somanath2020-of}.
We measure both the reconstruction time and visual quality for all the methods.
We use the Peak signal-to-noise ratio (PSNR) and Structural Similarity Index (SSIM) for quantitative visual quality comparison. 
The PSNR and SSIM values of each method are calculated by comparing the rendered virtual object to the physical object.
In this work, we use the classical physical mirror ball as it can be easily acquired. 
The higher the values of PSNR and SSIM, the better the visual performance. 

We use the simulator to evaluate \sysname's performance in a wider range of scenarios. 
Our simulator allows easy extraction of ground truth lighting information at any reconstruction position in a photorealistic 3D indoor scene.
For simulation-based evaluations, \sysname is evaluated with six objects of different shapes and materials and is compared to two baselines:
\1 using a 360$^\circ$ camera at the observation position, akin to \cite{tuceryan2019ar360};
and \2 \Xihe, a recent academic framework that produces real-time low-frequency lighting estimation from RGB-D images~\cite{xihe_mobisys2021}.
We describe the synthetic dataset used in our study in \S\ref{subsec:simulation_environment_setup}. 

To provide an in-depth evaluation of \sysname's performance, we also conduct a number of ablation studies that demonstrate the quality-performance trade-offs (\S\ref{subsec:eval_quality_perf_trade_offs}), highlight our design choices for near-field and far-field reconstructions, as well as identify applicable scenarios (\S\ref{subsec:eval_near_field_recon} and \ref{subsec:eval_far_field}). 
The three quality presets for near-field reconstruction are configured as following:
\1 \sysname (low): number of views is 3, color image resolution is 256x192, \dpcp resolutions are [512x256, 256x128, 64x32], environment map resolution is 512x256;
\2 \sysname (medium): number of views is 4, color image resolution is 512x384, \dpcp resolutions are [768x384, 384x192], environment map resolution is 512x256;
\3 \sysname (high): number of views is 5, color image resolution is 1024x768, \dpcp resolutions are [1024x512, 512x256], environment map resolution is 1024x512.
All three presets for far-field reconstruction have the color image resolution of 32x24 and share the same environment map resolution configurations as near-field reconstruction.

\subsection{Testbed-Based System Performance}

\begin{figure}[t]
\includegraphics[width=0.95\linewidth]{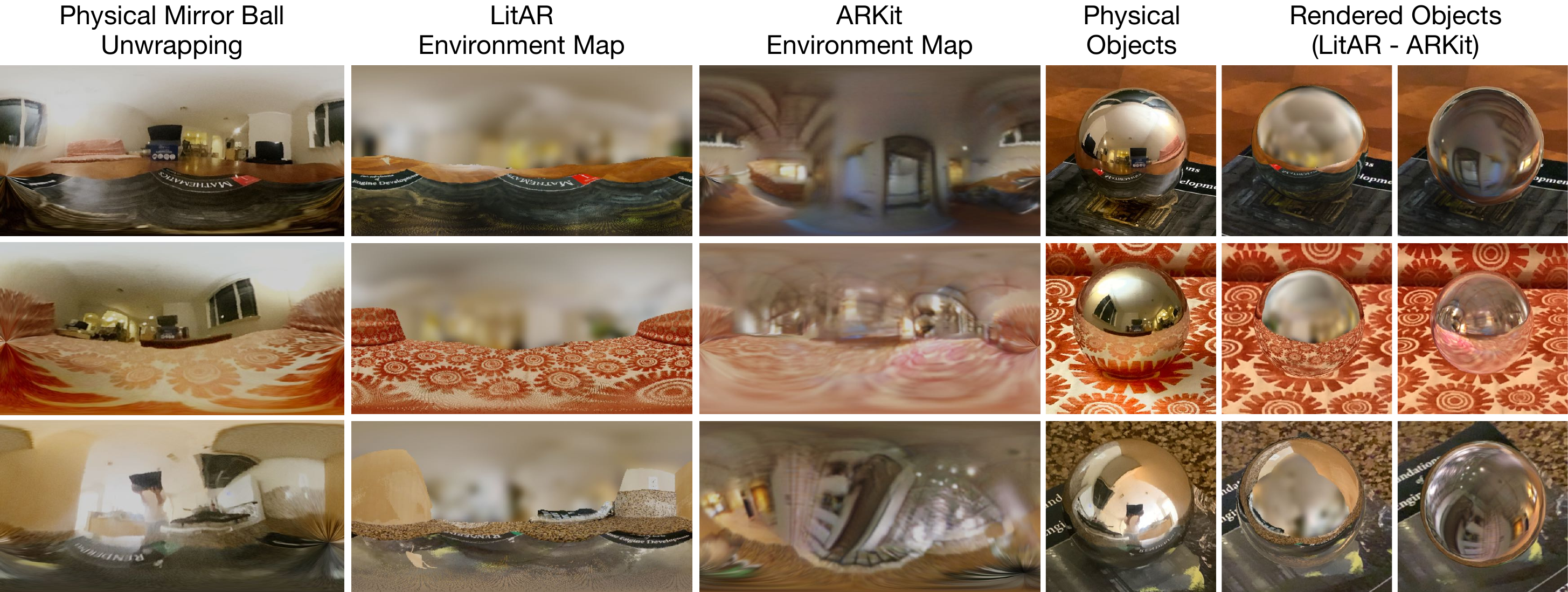}
\caption{
Qualitative comparison between \sysname and ARKit on three real-world indoor scenes.
    \textnormal{
        Each row represents an indoor scene.
        \emph{Column 1:} the panorama view of a scene at the reconstruction position by unwrapping the physical mirror ball reflection~\cite{debevec2006image}.
        \emph{Column 2:} \sysname's environment maps have good visual quality and rich details for the near-field portion while maintaining the structural similarity to the corresponding physical scene.
        \emph{Column 3:} ARKit's environment maps show varying performance, sometimes completely different from the scene (the first row), while others with less visual details. 
    }
  }
  \label{fig:end_to_end_sys_rendering}
\end{figure}

\begin{figure}[t]
    \centering
    \begin{subfigure}[b]{0.48\linewidth}
        \centering
        \includegraphics[width=\linewidth, trim=0px 25px 0px 3px,clip]{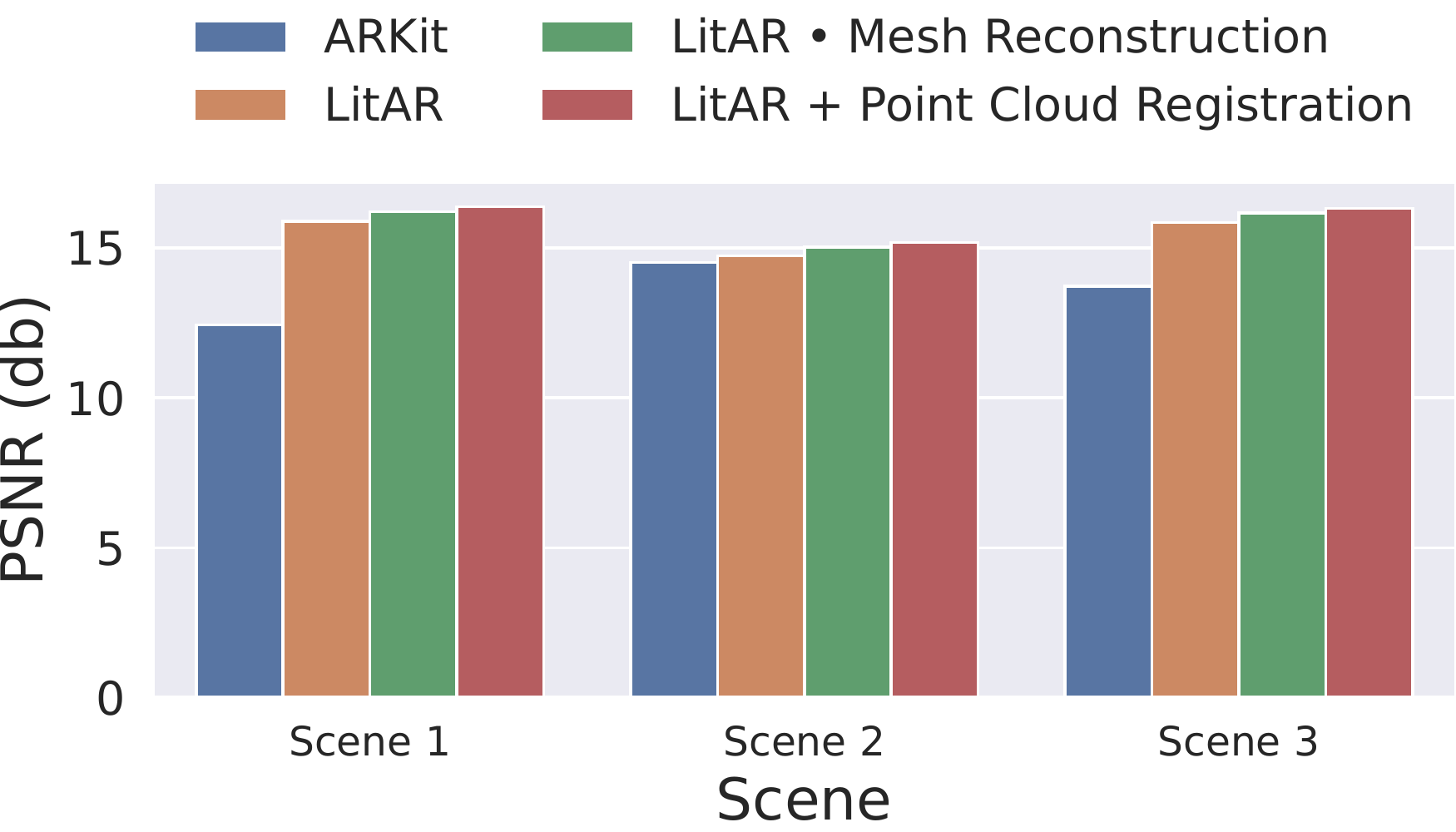}
        \caption{PSNR comparison}
        \label{subfig:visual_quality_psnr}
    \end{subfigure}\quad
    \begin{subfigure}[b]{0.48\linewidth}
        \centering
        \includegraphics[width=\linewidth, trim=0px 25px 0px 3px,clip]{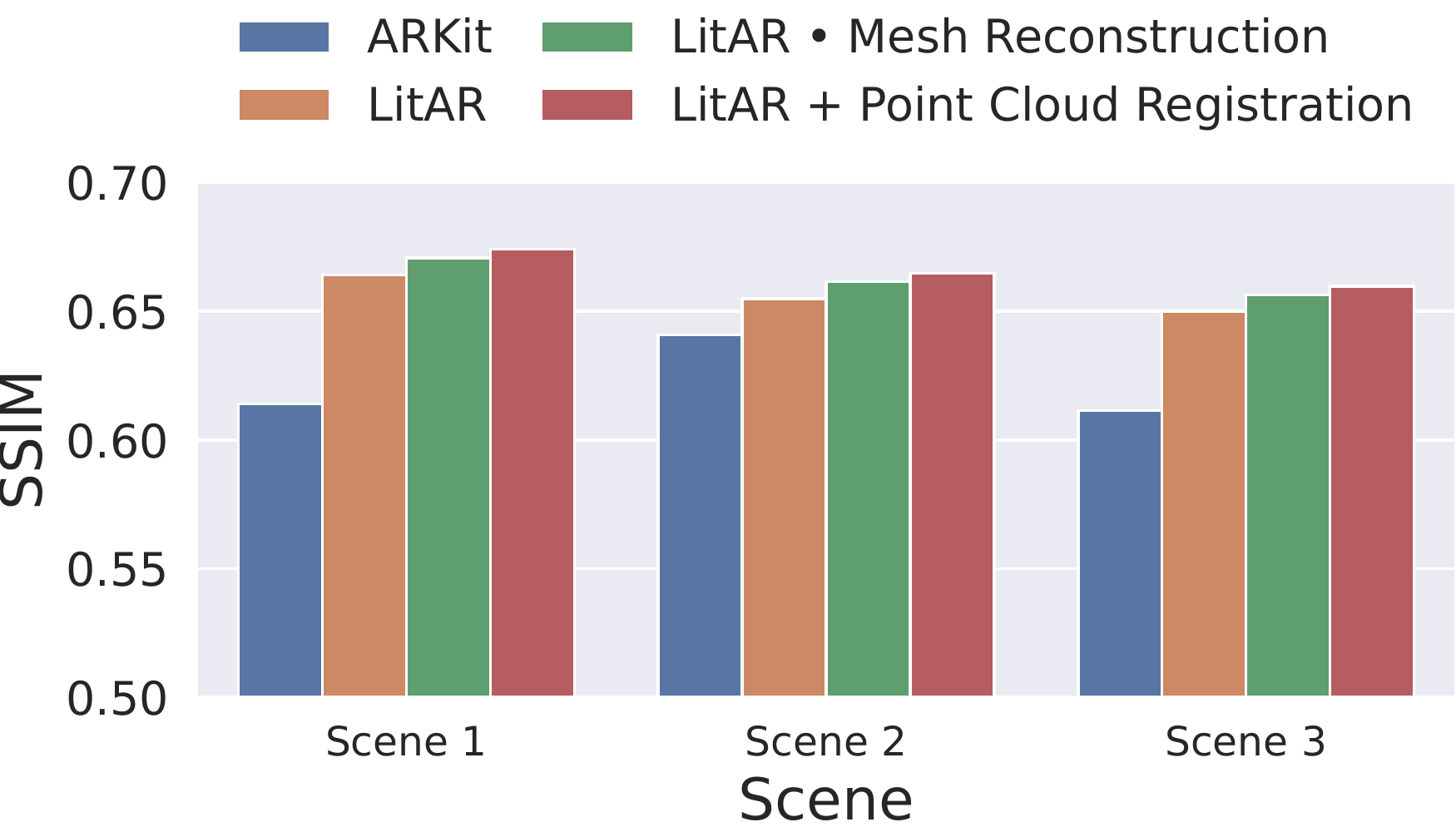}
        \caption{SSIM comparison}
        \label{subfig:visual_quality_ssim}
    \end{subfigure}
    \caption{
Quantitative comparison between \sysname and ARKit in terms of PSNR and SSIM.
        \textnormal{
        \sysname outperforms ARKit for all three real-world scenes.
        Further, using the conventional mesh reconstruction in place of our lightweight \dpcp only increases the PSNR/SSIM values slightly by 2\%/1\%. 
        Similarly, turning on the asynchronous point cloud registration has only minor improvement by 3\%/1.5\%. 
        Recall that mesh reconstruction and point cloud registration can take a few seconds (\S\ref{subsec:env_map_renderer}) and a couple hundred milliseconds (\S\ref{subsec:point_cloud_registration}), respectively.
        In short, \sysname achieves the best trade-off between visual quality and runtime performance.
        }
    }
  \label{fig:e2e_profiling}
\end{figure}

\subsubsection{End-to-end Evaluation.}

We compare the end-to-end rendering visual results and the runtime performance of \sysname and ARKit.
As shown in Figure~\ref{fig:end_to_end_sys_rendering} (last two columns), the virtual mirror balls rendered with \sysname have more reflection details and better color tune than those rendered with ARKit's learning-based method.
Specifically, for all three scenes, \sysname's virtual balls have higher visual similarity than physical mirror balls.
In contrast, the virtual balls rendered using ARKit either reflect incorrect indoor scenes (especially on the far-field portion of the environment) or lack fine-grained visual details, e.g., the text on the book cover. 
We can more easily observe such visual quality differences by comparing the generated environment maps (the second and third columns) to the unwrapped images from the physical mirror ball.
Note that the unwrapped images do not represent the ground truth lighting as they are often distorted but can serve as a visual guide of the panorama view at the reconstruction position.
By comparing the environment maps generated by \sysname to the unwrapped images, we see that \sysname can accurately reconstruct scene elements from near-field observations while faithfully recovering environmental geometry and color tone information from far-field observations. 

Figure~\ref{fig:e2e_profiling} quantifies the visual quality using two commonly used image metrics, i.e., PSNR and SSIM, by comparing the rendered object to a physical mirror ball image at the same reconstruction position.
We see that \sysname outperforms ARKit on all three real-world captured scenes, with up to 14.3\% higher PSNR and 5.5\% higher SSIM.
When replacing our lightweight \dpcp component of \sysname with the Ball-Pivoting surface mesh reconstruction~\cite{bernardini-ball-pivoting}, we only notice a minor increase in the PSNR/SSIM values.
This observation demonstrates the effectiveness of \dpcp for generating high-quality near-field reflections.
Similarly, we do not observe significant improvement when running \sysname with the point cloud registration component.
We suspect this is because AR frameworks such as ARKit can provide reasonable device tracking data in most cases with slow movement.
We observe that the tracking data of ARKit often drifts in cases of fast movement, making the point cloud registration component integral.
We omit their visual effect comparisons as both of the \sysname's variations do not show noticeable visual quality differences to \sysname. 

Finally, the average end-to-end latency of near-field and far-field reconstruction is 134.4 ms and 57.5 ms, respectively.
Detailed component-wise time breakdown is discussed in the next section.
These latencies translate to updating high-quality lighting roughly at 22 fps, i.e., every 134.4 ms \sysname can provide one near-field and two far-field environment maps.
Such update frequency should be sufficient for most AR applications~\cite{Xu2021-sn,Yi2020-na}.
For AR applications that require higher update frequency, we can either resort to more powerful edge servers (currently using an energy-efficient Jetson board) or use a lower quality setting, as discussed in the next section.

\subsubsection{Trade-Offs Between Rendering Quality and Runtime Performance }
\label{subsec:eval_quality_perf_trade_offs}

\begin{figure}[t]
    \begin{subfigure}[b]{\linewidth}
        \centering
        \includegraphics[width=0.95\linewidth]{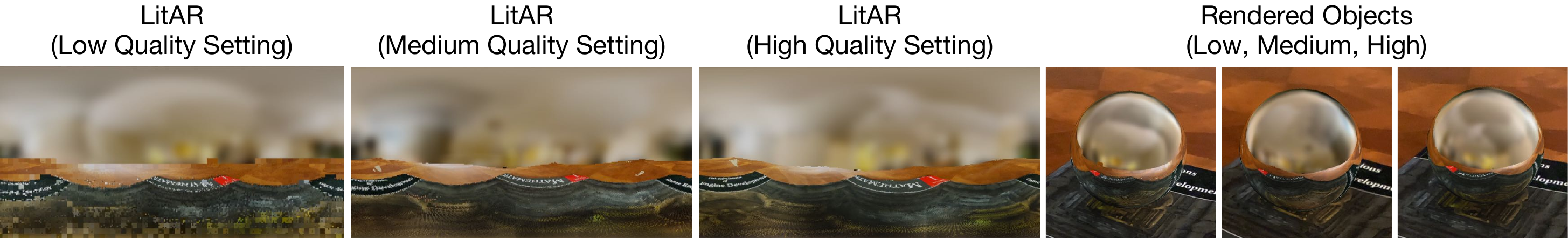}
        \caption{Rendering visualization on different quality presets}
        \label{subfig:preset_visualization}
    \end{subfigure}
    \vspace{0.1em}
    
    \centering
        \begin{minipage}{.29\linewidth}
        \begin{subfigure}[b]{\linewidth}
            \centering
\includegraphics[width=\linewidth]{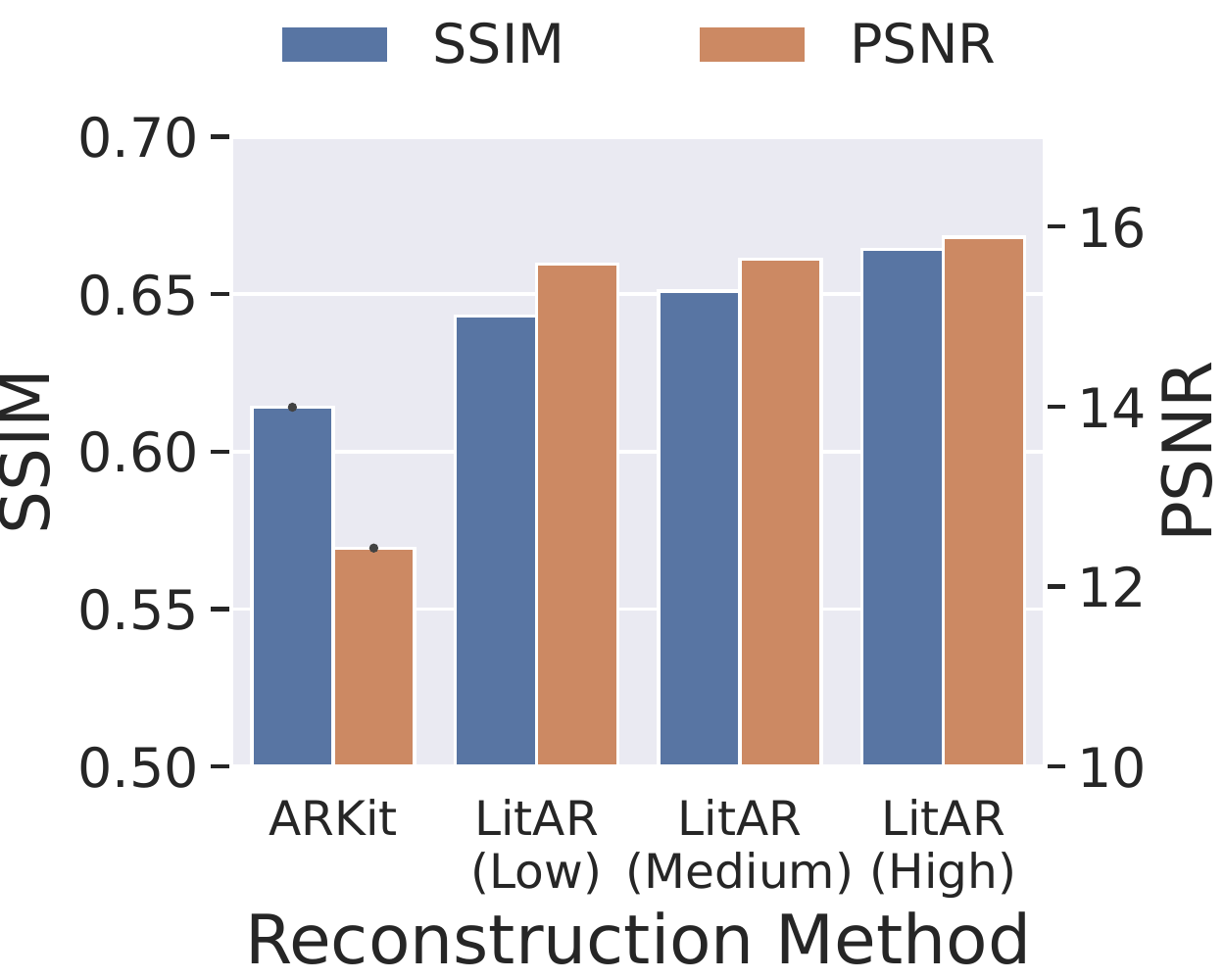}
            \caption{Visual quality comparison 
}
            \label{subfig:visual_quality_comparision}
        \end{subfigure}
    \end{minipage}
\quad
\begin{minipage}{.65\linewidth}
        \scriptsize \centering \begin{tabular}{lrrrrrr}
            \toprule
            \multirow{2}{*}{\textbf{System Component}} &
            \multicolumn{3}{c}{\textbf{Near-field (ms)}} &
            \multicolumn{3}{c}{\textbf{Far-field (ms)}} \\
            \cmidrule(lr){2-4}\cmidrule(lr){5-7}
                &
            low & Medium & High &
            low & Medium & High \\
\midrule
            
            Data Encode & 5.11 & 15.01 & 50.13 & 0.18 & 0.18 & 0.18 \\
            Data Offload & 4.07 & 7.78 & 18.24 & 0.55 & 0.55 & 0.55 \\
            
            Dense Point Cloud Generation & 8.39 & 9.71 & 15.62 & N/A & N/A & N/A \\
            Multi-resolution Projection & 9.02 & 9.89 & 14.56 & N/A & N/A & N/A \\
            
            Sparse Point Cloud Generation & N/A & N/A & N/A & 0.03 & 0.03 & 0.03 \\
            Anchor Extrapolation & N/A & N/A & N/A & 19.03 & 20.32 & 21.72 \\
            
            Environment Map Download & 20.01 & 29.11 & 35.83 & 20.02 & 29.71 & 35.03 \\

            \midrule
            End-to-end Reconstruction & 46.6 & 71.5 & 134.38 & 39.81 & 50.79 & 57.51 \\
            
            \bottomrule 
        \end{tabular}\vspace{0.5em}
        \subcaption{Reconstruction latency}                \label{tab:sys_profiling}
    \end{minipage}
    \caption{
        \sysname rendering quality-performance trade-offs.
        \textnormal{
            We show the time breakdown for \sysname to generate a near-field and far-field environment map.
            All quality settings achieve better SSIM than ARKit.
            A large portion of time, at least 26.46\%, was spent on edge-related operations (data encode/offload and environment map download).
            This observation suggests that \sysname has the promise to deliver high-quality environment maps directly on the device as mobile device GPU becomes more powerful.
        }
    }
    \label{fig:breakdown}
\end{figure}

We compare the rendering quality and latency of \sysname under different presets.
Figure~\ref{subfig:preset_visualization} shows the corresponding visual results.
We note that the environment maps generated at all three settings present visually coherent near-field reflection and correct anisotropic far-field color tones.
We can observe some pixelation effects in the environment map and the rendered mirror ball object for the low-quality preset due to low capturing resolution. 
All three settings achieve better quality than ARKit in terms of PSNR and SSIM values. 
For example, the low-quality preset has a 4.5\% higher SSIM value than ARKit.
Moreover, the difference in visual quality among the three presets is marginal, with only up to 3.5\% between low and high quality.

Furthermore, we measure the time breakdown of \sysname's near-field and far-field reconstructions. 
Table~\ref{tab:sys_profiling} shows the average performance over three runs.
For near-field reconstruction, the processing time of each component increases with the quality setting. 
For example, the time to encode the camera observations sees similar increases as the capturing resolutions, about 10X with 16X more pixels. 
We note that with the high-quality setting, the total time to encode and upload data takes 68.4 ms, about 1.9X of the environment map downloading time, even though the uploading/downloading resolution ratio is 1.5X.
In contrast, in the medium-quality setting, with the same uploading/downloading resolution ratio, it is 21.7\% faster to encode and upload data than to download.
This is because the device data is uploaded in the format of YCbCr 4:2:0, which has a smaller data size than the RGB environment map under the same resolution. 
Note that we are sending back the uncompressed RGB environment map for quality consideration. 
This observation suggests an interesting trade-off presented by the data encoding scheme in a real-world deployment. 
Moreover, this result also demonstrates that network-related operations (an artifact of using the GPU-based edge device) take up most of the end-to-end time, at 62.6\%, 72.6\%, and 77.5\% for low, medium, and high-quality settings, respectively.
The network performance bottleneck implies an immediate performance gain by directly using mobile GPU to run the entire reconstruction pipeline.

The far-field reconstruction presents a similar but slower upward increase in total time with the quality presets.
In particular, the time to decode/offload image data and generate a sparse point cloud is the same for all three quality presets. 
In all settings of far-field reconstruction, we downsample the native color images from 1920x1440 to a fixed resolution of 32x24.
The anchor extrapolation and environment map downloading time increase slightly with the environment map size.
To put the latency performance of \sysname in context, we note that a recent physical probe-based framework requires 30$\sim$400 ms to generate environment map~\cite{prakash2019gleam}. 
To understand the ARKit's performance, we use the inference time of its underlying deep learning model EnvMapNet~\cite{Somanath2020-of} since ARKit does not expose APIs to measure the end-to-end environment map generation.
Even though the EnvMapNet model can run in 9 ms, as reported by Somanath et al.~\cite{Somanath2020-of}, we have shown previously that it often leads to inferior visual quality compared to \sysname.
Moreover, the total time for ARKit to generate an environment map is likely to be similar to \sysname if accounting for other necessary steps, including data capturing and memory copying.
In short, this detailed breakdown analysis demonstrates that far-field reconstruction can achieve real-time performance for all presets; when used in conjunction with near-field reconstruction, the mobile AR applications can receive a sufficient number of environment map updates per second.

\subsubsection{Impacts of User Movement and Dynamic Scene}

We investigate the impact of user movement on the rendering quality of \sysname. Specifically, we are interested in understanding the need for our motion-based automatic capturing policy. One of the authors (referred to as the participant) used the \sysname-powered AR app on the iPad to perform a virtual object placement following a pre-determined trajectory. The participant was instructed to keep the same distance to the placement position and only to move the iPad around the placement position, resulting in a semi-circular trajectory. Further, the participant was asked to pause the movement every 30 degrees and to keep the camera views centered at the placement position. 
The experiment was repeated with three timer values, i.e., disabled, 300 ms, and 500 ms, which control the frequency to check the movement. 
First, we observe that our motion-based automatic capturing policy successfully detected the movement and resulted in six captured views (captured when the participant was static). Second, when comparing the generated environment map to the ground truth environment map captured by the mirror ball (at the same placement position), we did not observe noticeable visual quality differences for all three timer values. For example, the PSNR values for the timer=$300$ ms stayed relatively the same for the entire experiment, at 13.37 db ($\pm$ 0.25 db). Our observations suggest that \sysname can provide visually coherent renderings under user movement when the physical scene is static. Additionally, increasing the number of views (i.e., from one to six) provides limited visual quality improvement. This is intuitive as most near-field observations can be captured in a single view when the environment is simple and static.

In a second experiment, we created a simple dynamic scene by manually moving the physical object within the near-field observations. Specifically, the participant was asked to fix the iPad's position and select the placement position on a math book (similar to Figure~\ref{fig:teaser}). While using the AR app, the participant moved the book in various directions. We observed that \sysname could update the virtual object reflection to present details of different parts of the book. Even though the lighting reconstruction task does not block the rendering task, we still observed slightly choppy reflections. Please refer to the accompanying video for the visual quality demonstration. Two key factors impact the choppiness: \1 the physical scene change rate and \2 the reconstruction time. If the physical scene changes very rapidly (e.g., faster than the reconstruction time), the virtual object reflection will be perceived to lag. In addition to further speed up the lighting reconstruction, we suspect techniques that smooth the transition between two distinct environment maps (e.g., image fade in) and policies that pipeline the lighting reconstruction requests to mask network latency can also improve the user-perceived performance. We leave such investigations as future work.

\subsection{Simulation-Based Performance Evaluation}
\label{subsec:end2end}

\subsubsection{Synthetic Dataset Generation}
\label{subsec:simulation_environment_setup}

We describe the methodology we followed to generate a synthetic dataset using the Unity-based simulator (see Figure~\ref{fig:simulation_exp_workflow}). 
In a synthetic indoor scene, we first manually choose ten reasonable positions to be considered as lighting reconstruction positions for placing virtual objects.
Example reconstruction positions include on the floor or table. 
We vary several factors for each reconstruction position, including the number of capturing positions, mobile user/device height, and observation distance, to generate 72 camera observations.
Specifically, we set up a circular capturing trajectory with eight positions by evenly dividing the trajectory.
We decide the height and radius of the capturing trajectory by simulating possible scenarios when the mobile user is holding the device at chest height from a reasonable distance to the reconstruction position.  
We choose three typical human height values at $\{160, 170, 180\}$ centimeters and calculate the height of the trajectory by multiplying the user's height by 0.8~\cite{standard_proportions_of_the_human_body}.
We further measure the radius of the trajectory using the number of steps and choose three possible values of $\{0.5, 1, 1.5\}$ steps and use the height multiplied by 0.3 as the step length~\cite{stepping_science}.
For each camera observation, we export the camera HDR observation image, depth image, position, orientation, and ground truth lighting in the format of an equirectangular panorama image.

\begin{figure}[t]
    \centering
    \begin{subfigure}[b]{0.45\linewidth}
        \centering
        \includegraphics[width=\linewidth]{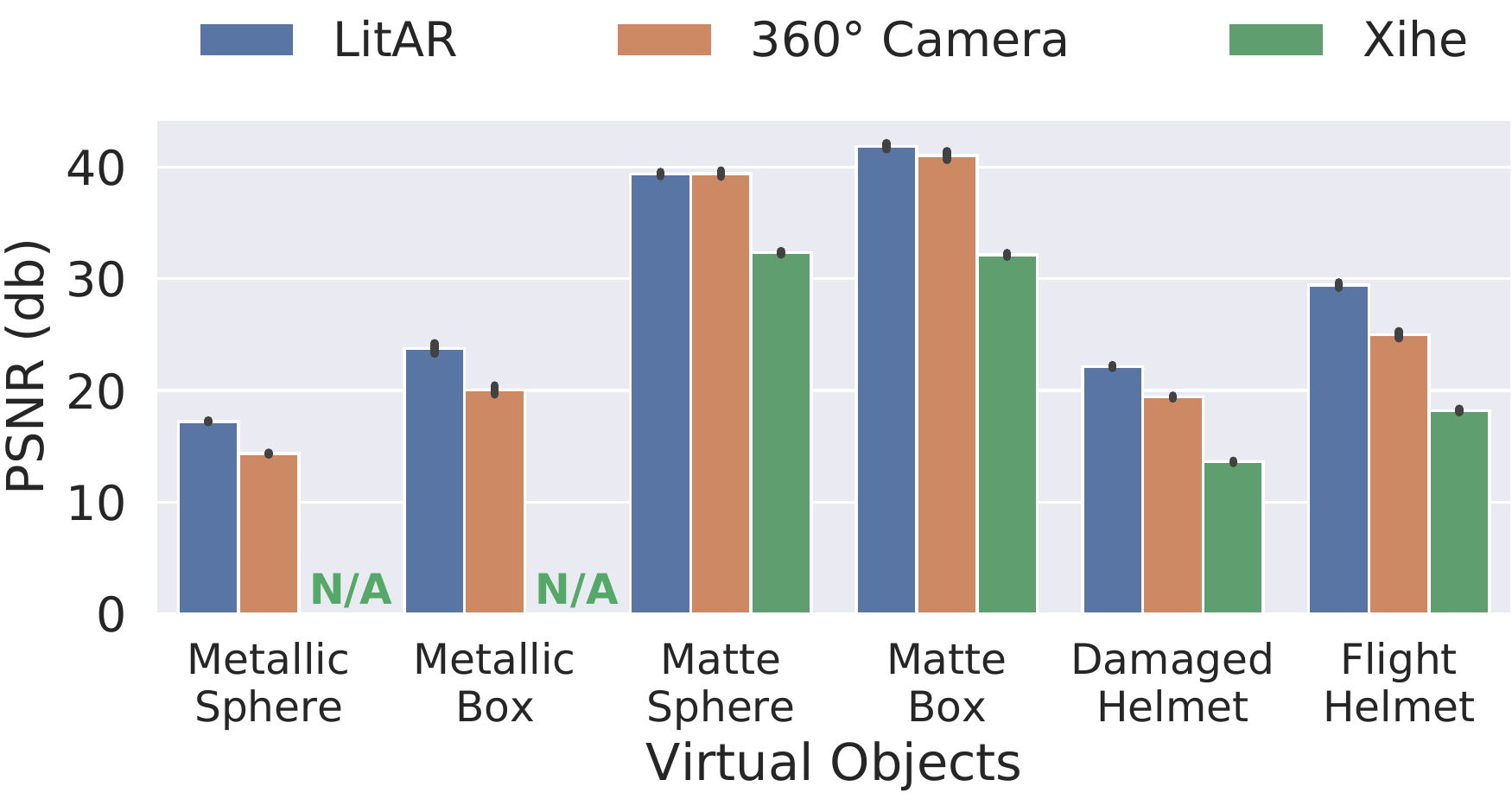}
        \caption{PSNR comparison}
        \label{subfig:sim_end2end}
    \end{subfigure}\quad
    \begin{subfigure}[b]{0.45\linewidth}
        \centering
        \includegraphics[width=\linewidth]{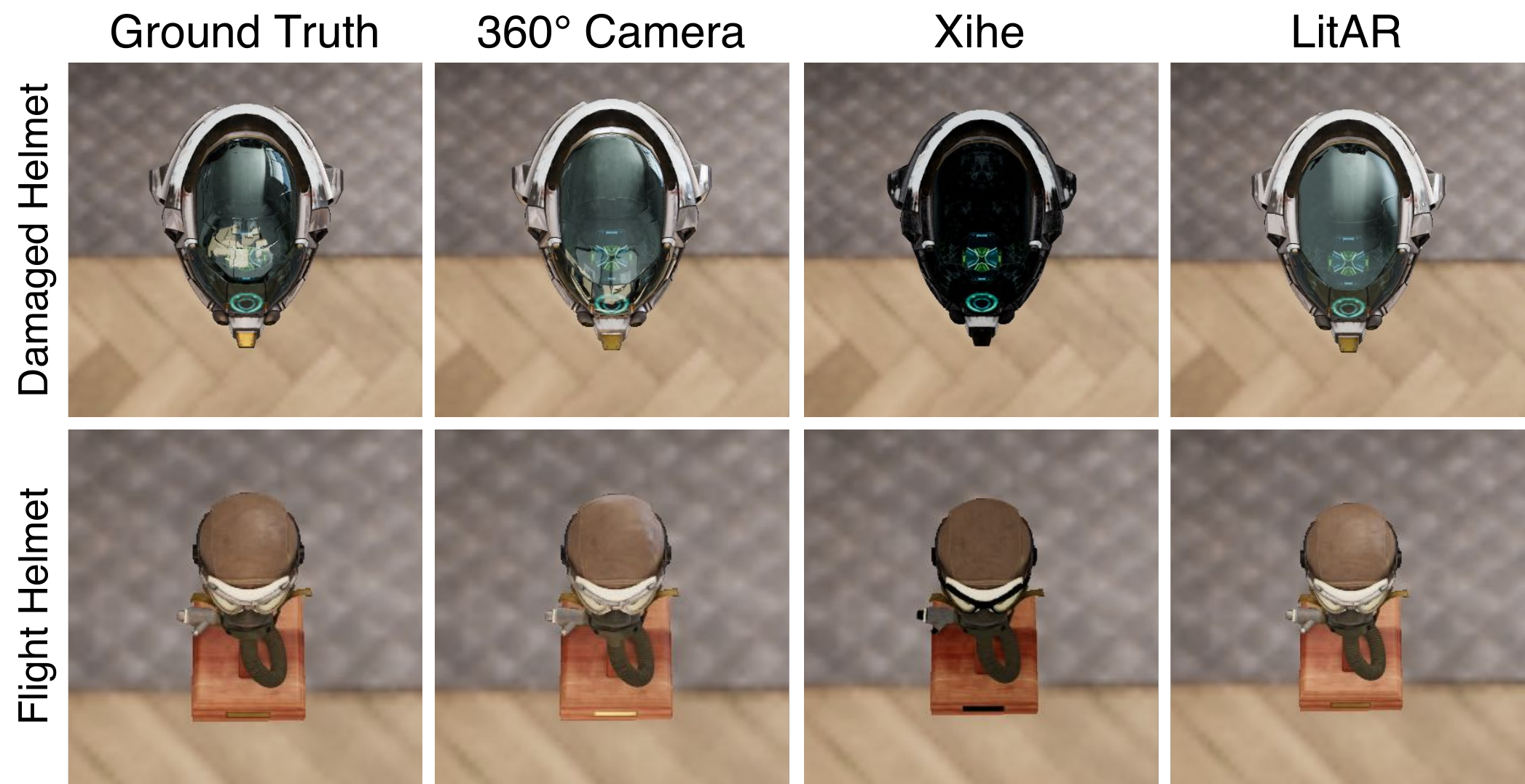}
        \caption{Rendering visual comparison}
        \label{subfig:sim_rendering}
    \end{subfigure}
  
    \caption{
        Simulator-based visual quality comparison.
        \textnormal{
            \sysname achieves better rendering effects, i.e., higher PSNR values (calculated against ground truth rendering), than other techniques for all objects. 
            The PSNR comparisons to \Xihe are omitted for the first four virtual objects as \Xihe only provides spatial-variant low-frequency lighting estimation~\cite{xihe_mobisys2021}.
            Virtual objects are rendered with lighting provided by ground truth lighting, 360$^\circ$ camera, \Xihe~\cite{xihe_mobisys2021}, and \sysname.
        }
  }
  \label{fig:end_to_end}
\end{figure}

\subsubsection{End-to-end Visual Quality Comparison}

We compare the end-to-end rendering performance quantitatively and qualitatively on six different virtual objects.
For this experiment, we configure the simulator to run the \reconstruction to process one near-field observation and nine far-field observations based on the guided movement policy. 
For near-field reconstruction, we use mesh reconstruction instead of \dpcp to support the high-quality point projection.  
The following results showcase the upper bound of visual quality that \sysname can achieve.

Figure~\ref{fig:end_to_end} shows the comparisons of PSNR values.
Specifically, on complex objects with physically-based materials (i.e., \emph{Damaged Helmet} and \emph{Flight Helmet}), \sysname achieves 44.1\% and 12.1\% higher values of PSNR than a recent deep learning-based AR lighting estimation system~\cite{xihe_mobisys2021} and the lighting information captured by a 360$^{\circ}$ camera, respectively. 
This result indicates that by correctly leveraging user movement and scene geometry information, \sysname can generate highly accurate lighting from limited camera observations.
In addition, as we will see in Figure~\ref{fig:near_fields_rendering_alblation}, the rendering performance of \sysname is roughly the same with fewer observations.
Note that we omit the comparison to the rendering PSNR by \Xihe on metallic objects (\emph{Metallic Sphere} and \emph{Metallic Box}) because \Xihe only provides low-frequency lighting in \SHc format, which does not support reflective rendering.

Figure~\ref{subfig:sim_rendering} compares the visual effect of objects rendered with different lighting information. 
We observe that \sysname produces visually coherent virtual objects.
This observation suggests \sysname is effective in generating a complete high-quality environment map. 
Compared to those rendered with the 360$^{\circ}$ camera observations, helmets rendered with \sysname exhibit higher structural similarity to the ones rendered with ground truth lighting. 
For example, the two reflective regions in the lower bottom of the \emph{Damaged Helmet} are visually separated.

\subsubsection{Ablation Study of Near-Field Lighting Reconstruction}
\label{subsec:eval_near_field_recon}

\begin{figure}[t]
    \includegraphics[width=0.83\linewidth]{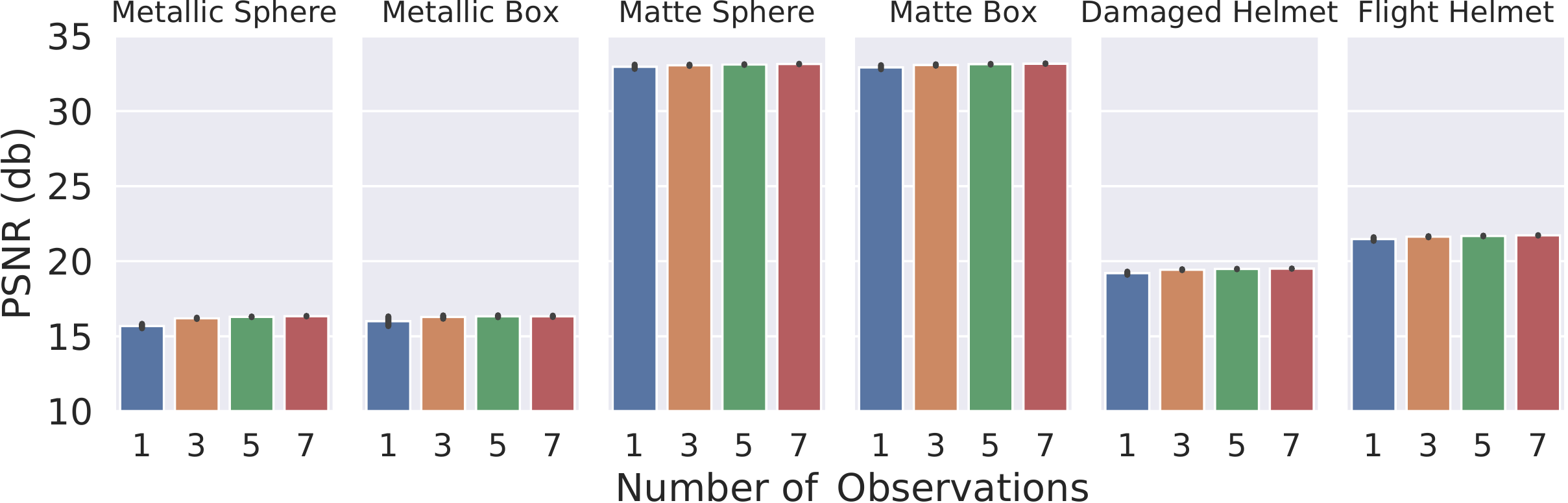}
    \smallskip

    \includegraphics[width=0.83\linewidth]{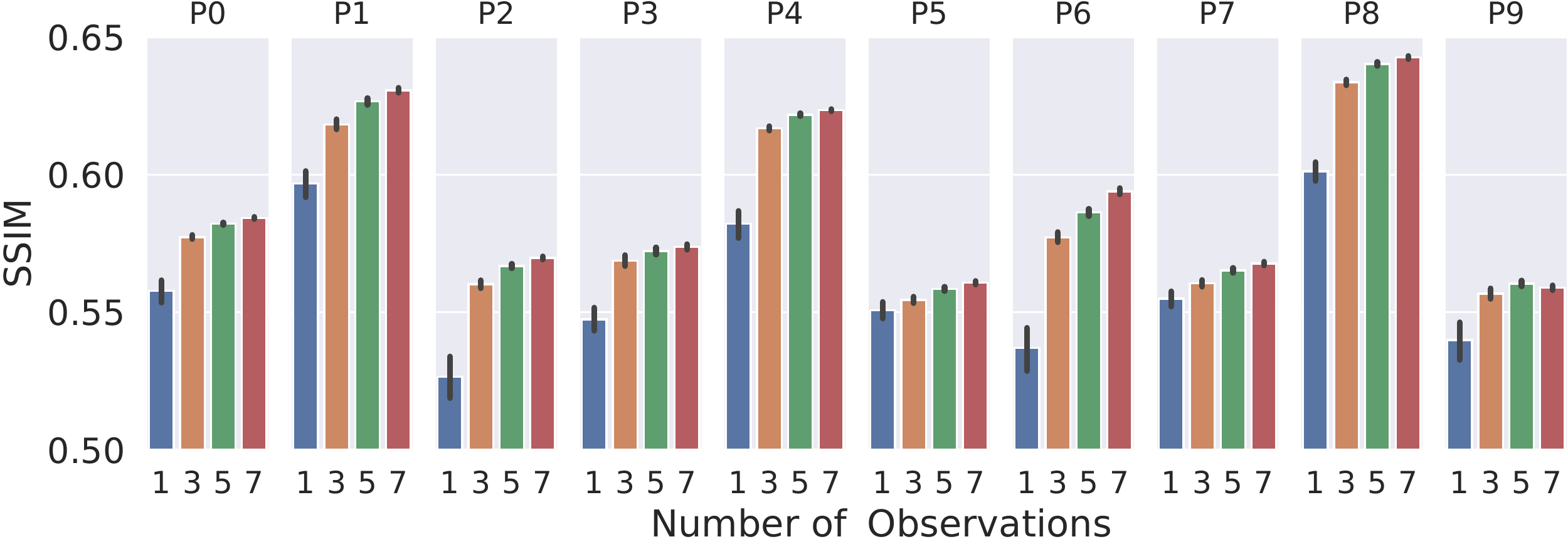}
    \caption{
    Quantitative comparison of rendering quality with different numbers of camera observations.
        \textnormal{
            The performance of \sysname, measured in PSNR, only exhibits slight improvement with more observations, while the SSIM values increase significantly for reflective materials.
        }
    }
    \label{fig:near_fields_rendering_alblation}
\end{figure}

This section demonstrates the effectiveness of our \reconstruction for the near-field observation and highlights the importance of \sysname's far-field design.
As we have designed \sysname to progressively improve the intermediate point cloud by naturally exploiting mobile user/device movement, we evaluate the impact of the number of near-field observations on the rendering results
We set up the experiment using our simulator as follows:
\1 for each observation position, we combine camera observations from ${3, 5, 7}$ nearby positions on the orbiting trajectory;
\2 we use mesh reconstruction with \sysname for combined observations.
To eliminate the impact of far-field observations, we use a single dominant image color to fill the far-field portion of the environment map to simulate the ambient light sensor data.
Figure~\ref{fig:near_fields_rendering_alblation} compares the rendering accuracy for different numbers of observations. 
We observe that the rendering PSNR values only increase \emph{slightly} with the number of observations. 
This observation suggests that only a small portion of the environment map needs to be processed with depth information, further motivating our design choice of reconstructing near-field and far-field observations separately.
Furthermore, we show that rendering SSIM values increase significantly, 0.057 on average across the tested positions in Figure~\ref{fig:near_fields_rendering_alblation}, with the number of observations for the \emph{Metallic Sphere} object for all ten tested reconstruction positions. 
This result is intuitive since more complete near-field reflections will improve the structural similarity. 
However, higher SSIM values do not always guarantee better PSNR values, thus implying the necessity to use both metrics in quantitative studies for lighting reconstruction.

\subsubsection{Ablation Study of Far-Field Lighting Reconstruction}
\label{subsec:eval_far_field}

\begin{figure}[t]
\includegraphics[width=0.82\linewidth]{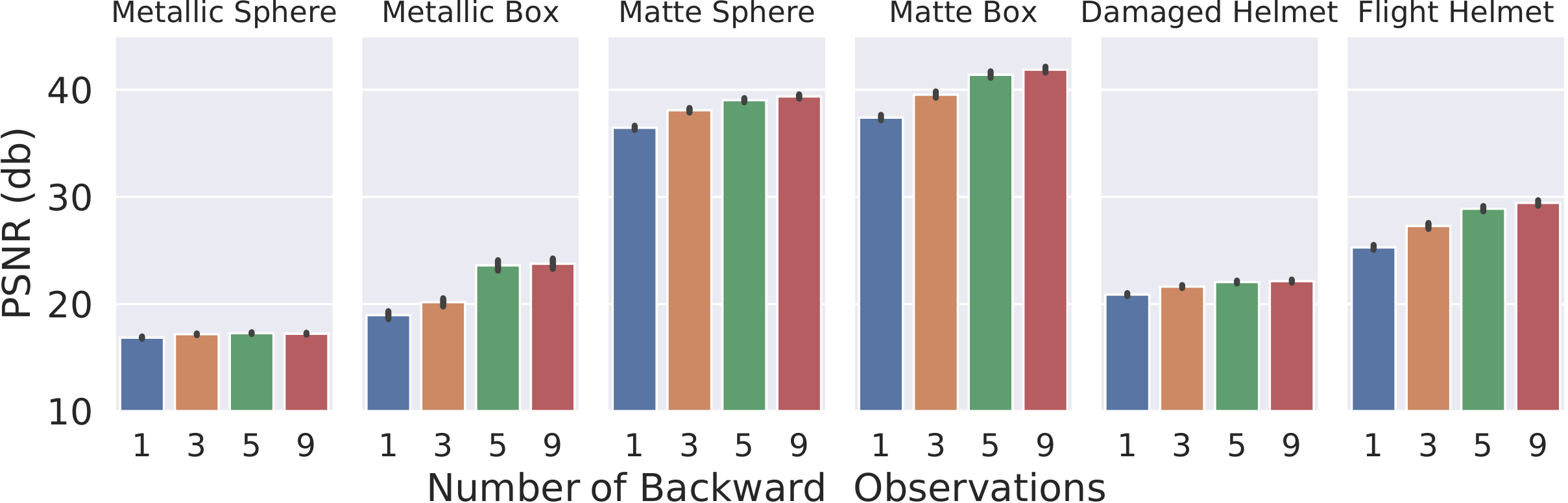}
\caption{
    Quantitative comparison of rendering quality for guided movements.
        \textnormal{
        Increasing the number of backward far-field observations have different levels of improvement for tested objects.
    }
  }
  \label{fig:guided_movement_rendering}
  \vspace{-3mm}
\end{figure}

So far, we have demonstrated the effectiveness of \sysname and its spatial variance-aware near-field reconstruction component. 
In this section, we evaluate the performance of \sysname's directional-aware far-field lighting reconstruction and the effectiveness of our guided movement policy.
We evaluate the rendering performance with different numbers of guided far-field observations.
Recall that guided movements naturally increase the observed scene, allowing \sysname to extrapolate environment map pixel color closer to the ground truth. 
Figure~\ref{fig:guided_movement_rendering} shows the comparison of rendering performance. 
We observe that for all tested objects, having access to more guided observations improves the PSNR value by up to 31.04\%. 
Furthermore, far-field observations present different levels of impact on objects with different shapes. 
For example, both box-shaped objects are improved more significantly than spherical objects.
This finding, if generalized, can help further improve usability by providing guided movements for different objects.

\section{Related Work}
\label{sec:related}

\subsection{Mobile-specific Lighting Support}
As mobile device capability increases and AR re-emerges in user-facing applications~\cite{Amazon_undated-vp, Inter_IKEA_Systems_B_V2017-iv}, obtaining environment lighting for photorealistic rendering has garnered increasing interests in the research communities~\cite{ChengSCDZ18_graph,prakash2019gleam,pointar_eccv2020,Somanath2020-of,xihe_mobisys2021}.
For brevity, we only discuss techniques targeted at lighting for indoor scenes.
On the more theoretical front, Cheng et al. leveraged both the rear and front cameras to estimate \shc~\cite{ChengSCDZ18_graph}, which is a low-frequency lighting representation that does not support reflection.
Zhao et al. proposed a two-staged pipeline called PointAR that leverages the mobile depth sensor to estimate spatially-variant lighting~\cite{pointar_eccv2020}.
Somanath et al. introduced an efficient deep learning (DL) model called EnvMapNet that generates HDR environment maps from LDR images~\cite{Somanath2020-of}.
In contrast to previous learning-based approaches, our work \sysname directly generates high-quality environment maps with the core technique of \reconstruction and several practical optimizations, including \dpcp.
As such, \sysname is not subject to common limitations of DL-based methods such as training data availability and inference performance on heterogeneous mobile resources.
We also note that \sysname includes a simulator that leverages 3D indoor scenes to conduct controlled experiments, thus avoiding the need for an expensive manual process of obtaining ground truth lighting information.

\subsection{System Supports for Lighting}
On the system front, commercial SDKs such as ARKit~\cite{arkit} and ARCore~\cite{arcore} provide easy-to-use lighting estimation APIs for mobile AR application development. 
Two recent academic frameworks improved upon commercial solutions: GLEAM provides a real-time mobile illumination framework that supports reflective virtual objects with the use of physical probes~\cite{prakash2019gleam}; \Xihe introduced a 3D-vision based framework that provides adaptive lighting estimation~\cite{xihe_mobisys2021}.
We design \sysname from the outset by considering mobile characteristics, including limited FoVs, natural device/user movements, and leveraging edge GPU assistance, which well positions it for high-quality and efficient reconstruction of environment maps.

\subsection{Image-based Lighting}
In addition to methods for mobile-specific lighting discussed above, many researchers have investigated image-based lighting~\cite{debevec2006image,corsini2008stereo,karis2013real}. For example, numerous works designed approaches for generating environment lighting representations from camera videos~\cite{havran2005interactive,unger2013temporally,grosch2007consistent}, and assisted lighting reconstruction with physical probes~\cite{debevec2008rendering}, object cues~\cite{Sun2019-rw}, or scene geometry~\cite{Maier2017-qp,Azinovic2019-nz}. 
Recent work is DL-based primarily and can broadly fall into two types based on the output, i.e., estimating low-frequency lighting~\cite{Garon2019,srinivasan20lighthouse} or high-frequency lighting~\cite{Gardner2017,Song2019}. 
For example, Gardner et al. proposed to divide the HDR environment map learning task into two subtasks and generated one lighting estimation result per image~\cite{Gardner2017}. 
Even though this work can handle the rendering of specular objects, it does not consider spatial variance.
On the contrary, both Garon et al.~\cite{Garon2019} and Lighthouse~\cite{srinivasan20lighthouse} support spatially-variant lighting but are limited in rendering reflective materials. 
Our work falls in between these two types of work by generating an environment map that consists of near-field and far-field components.
This hybrid environment map allows more effective reconstruction within mobile constraints such as user movement and depth-sensing accuracy while still achieving visually coherent rendering for various objects, including reflective ones.

\section{Conclusion}

In this work, we introduced an end-to-end lighting reconstruction system called \sysname that generates high-quality environment maps for mobile AR applications.
As quantitatively and qualitatively demonstrated, AR applications can use environment maps reconstructed by \sysname to render objects of various properties, including reflective materials, with 14.3\%/5.5\% higher PSNR/SSIM and better visual coherence than ARKit.
We showed that \sysname could produce virtual objects with more realistic and visually coherent reflection, as well as fine-grained visual details. 
We used physical object images for testbed-based experiments to serve as the basis of desired visual quality.
Furthermore, using our simulator, we compared against other techniques, including Xihe and 360$^\circ$ camera, by having access to ground truth lighting. 
We have released our research artifacts at \url{https://github.com/cake-lab/LitAR} to facilitate future research work in our community. 

Aside from the realistic and visually coherent rendering goal, we designed \sysname with mobile-specific constraints, e.g., limited sensing and data noise, in mind. 
By exploring mobile user behaviors and working within mobile sensing constraints, we proposed the \reconstruction scheme that divides camera observations into near-field and far-field observations based on pixels' relative distance to the reconstruction position.
\sysname can work with as few as one camera observation and can progressively improve the quality of generated environment maps, especially for metallic objects, with more camera observations.
Keeping usability in mind, we further introduced the motion-based automatic capture and guided bootstrapped movement policies to help AR users capture higher quality data and more suitable camera observations. 
\sysname significantly speeds up both the near-field and far-field reconstructions by two novel point cloud techniques, i.e., \dpcp and \extrapolate.
Last but not least, \sysname provides three quality presets and exposes several knobs for mobile AR applications to make reconstruction quality and time trade-offs based on their specific use cases.

We evaluated \sysname's performance with a lab-based testbed and a game engine-based simulator.
We observed that \sysname could generate higher-quality environment maps than ARKit and result in rendered objects with up to 14.3\%/5.5\% higher PSNR/SSIM compared to the physical counterpart.
Furthermore, we showed that \dpcp significantly reduces the point cloud projection from 3 seconds (using mesh reconstruction) to 14.6ms. 
Overall, \sysname can generate about 22 high-quality environment maps per second when point cloud registration is not required.
As we design the point cloud registration to run asynchronously, the registration step will not block the main reconstruction pipeline; once completed, \sysname will send an updated environment map to the mobile device.
As part of the future work, we will explore techniques to improve the details of the generated environment maps and design runtime policies to handle temporally variant lighting more robustly.
 
\begin{acks}
We thank the anonymous reviewers for their constructive reviews. 
This work is partly supported by NSF Grants CNS-1815619, NGSDI-2105564, and VMWare.
\end{acks}

\balance{
\bibliographystyle{ACM-Reference-Format}
\bibliography{main}
}

\end{document}


\maketitle

\begin{figure}
  \includegraphics[width=0.95\linewidth]{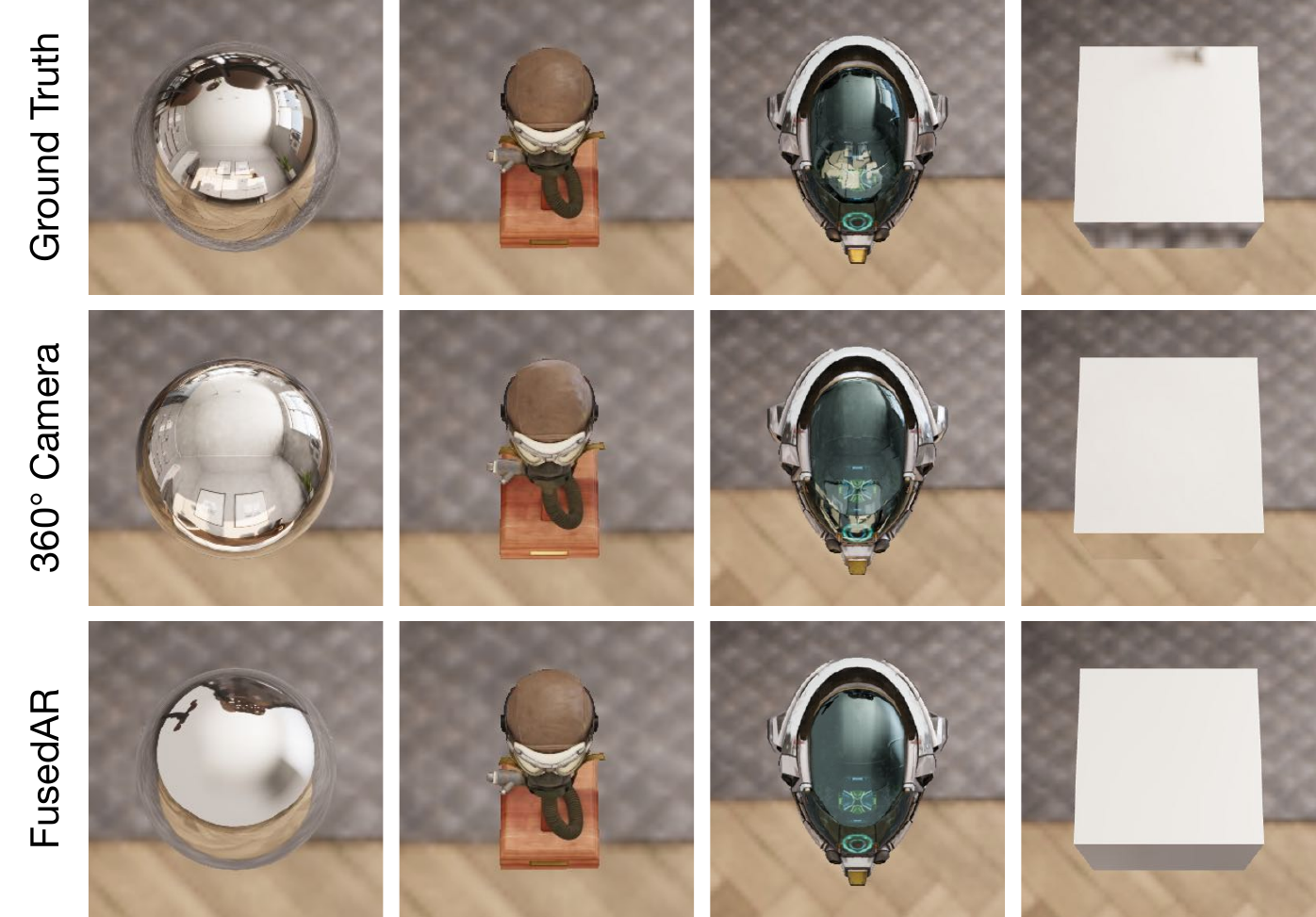}
  \caption{Full rendering comparison of the first row in teaser.
  }
  \label{fig:render_1_1}

\end{figure}

\begin{figure}
    \vspace{1em}
  \includegraphics[width=0.95\linewidth]{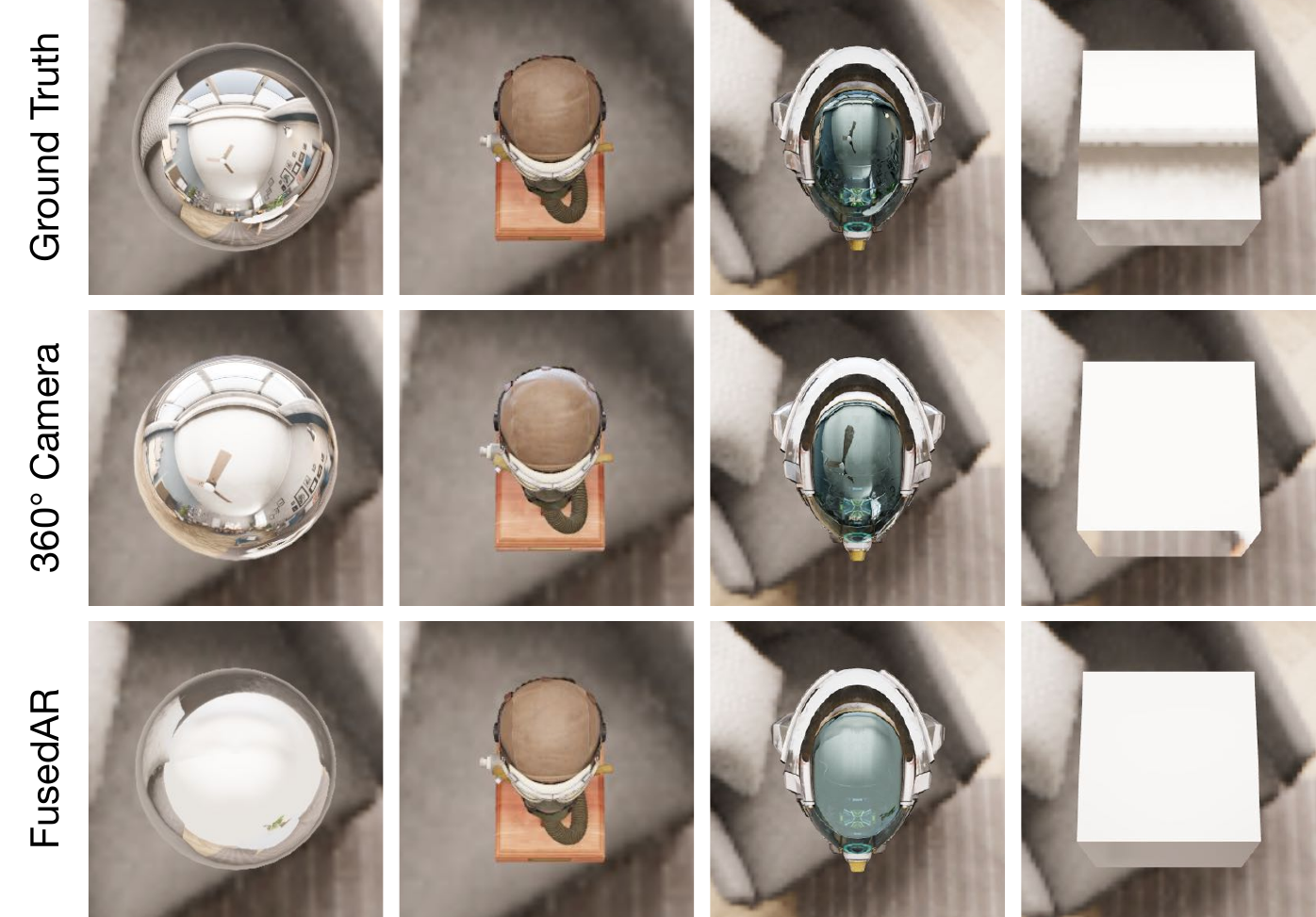}
  \caption{Full rendering comparison of the second row in teaser.
  }
    \label{fig:render_1_2}
\end{figure}

\begin{figure}
    \vspace{-6em}
  \includegraphics[width=0.95\linewidth]{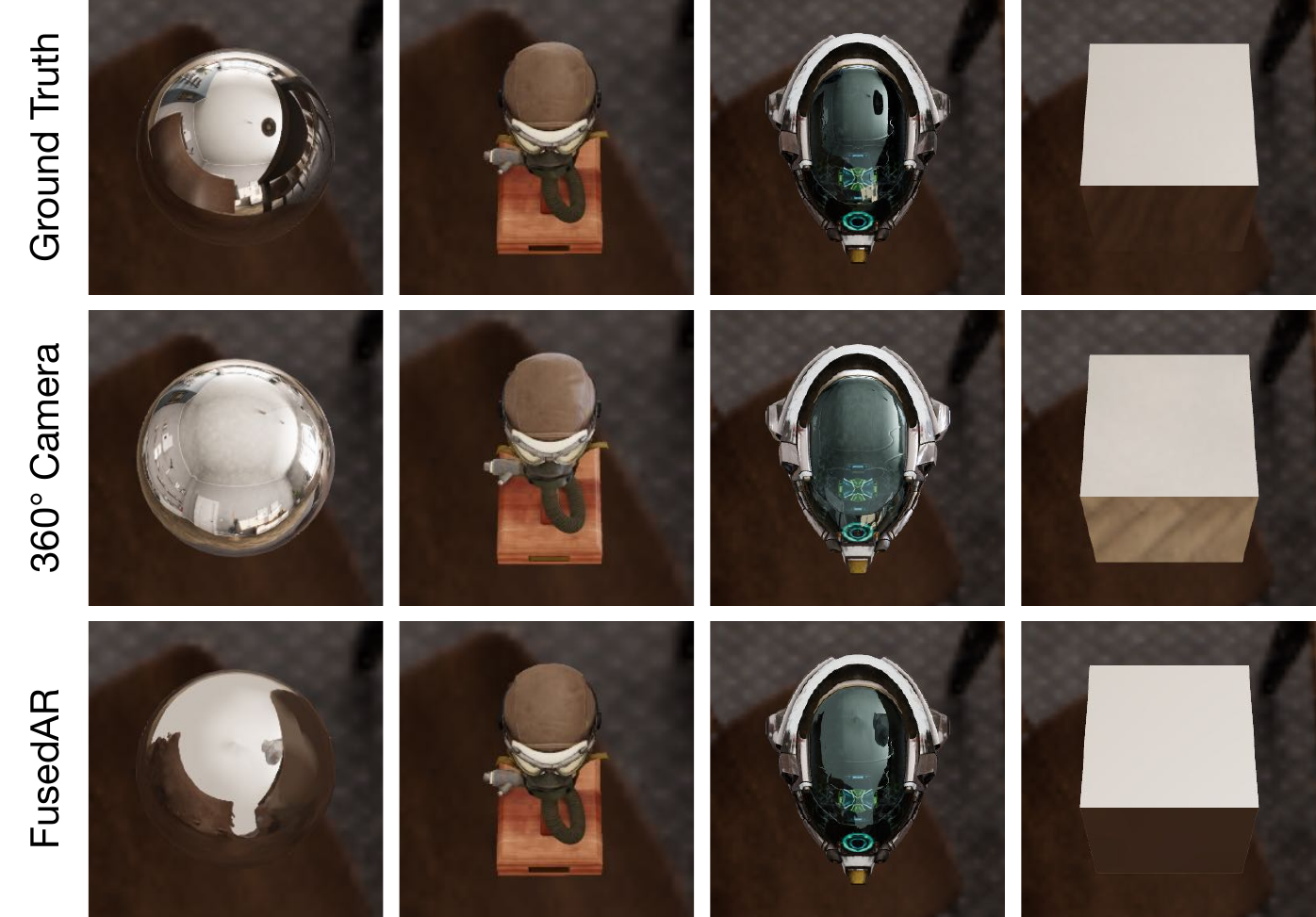}
  
  \caption{Full rendering comparison of the third row in teaser.
  }
    \label{fig:render_1_3}
\end{figure}

\begin{figure}
\vspace{-10em}
  \includegraphics[width=0.95\linewidth]{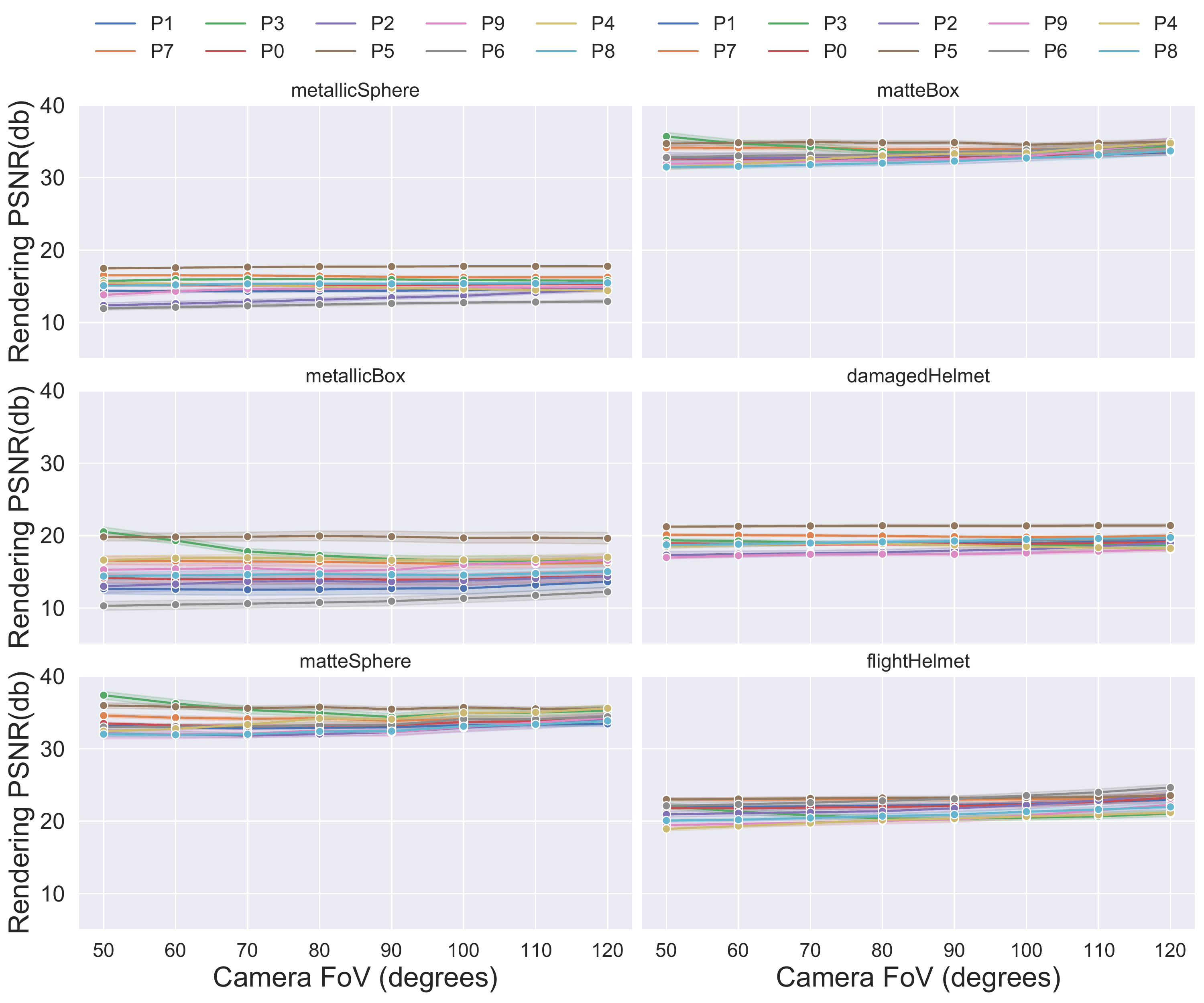}
  \caption{Detailed camera FoV impact on rendering accuracy.
  }
  \vspace{12em}
    \label{fig:render_1_3}
\end{figure}

\begin{figure}

  \includegraphics[width=0.95\linewidth]{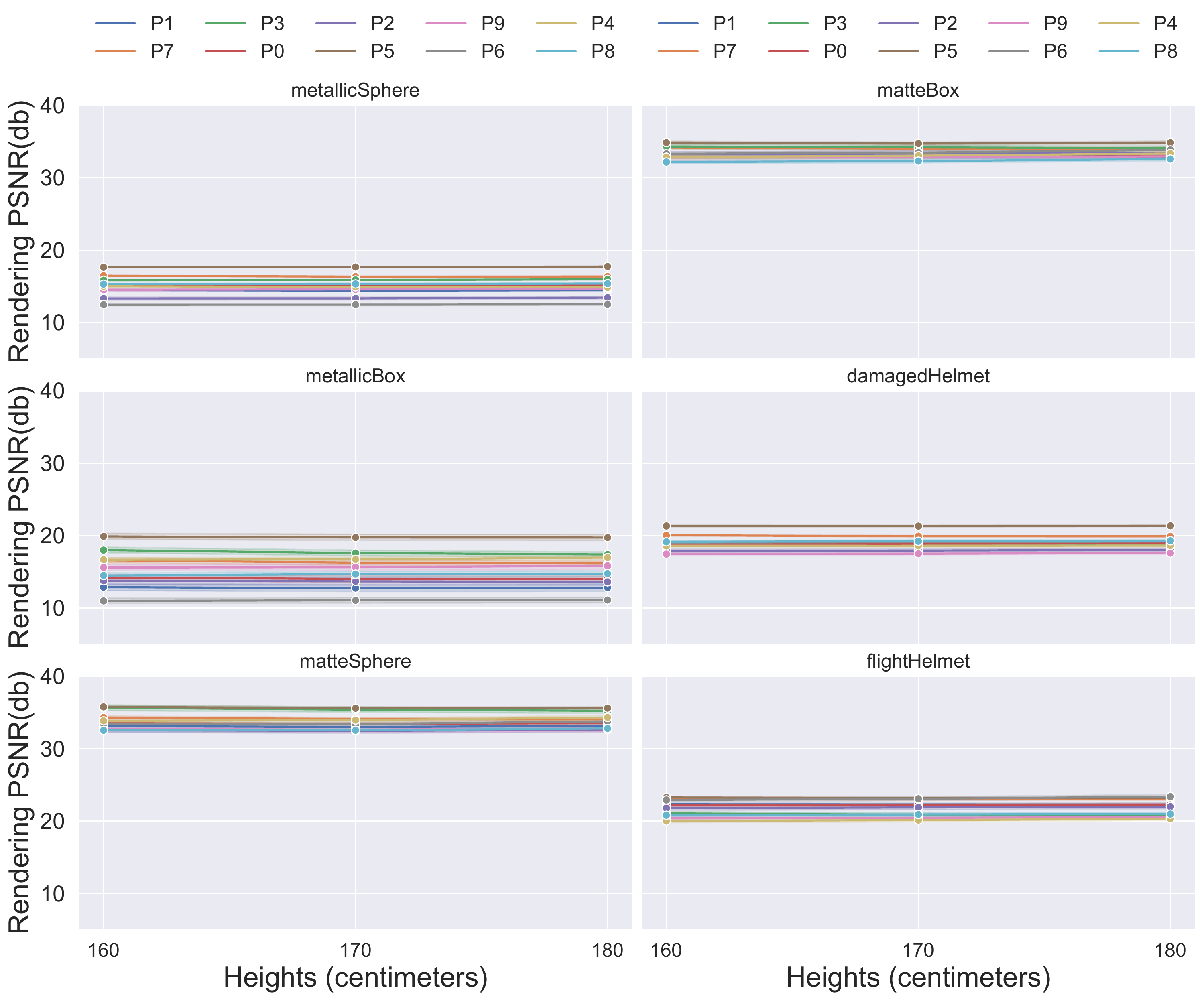}
  \caption{Detailed observer height impact on rendering accuracy.
  }
  
    \label{fig:render_1_3}
\end{figure}

\begin{figure}

  \includegraphics[width=0.95\linewidth]{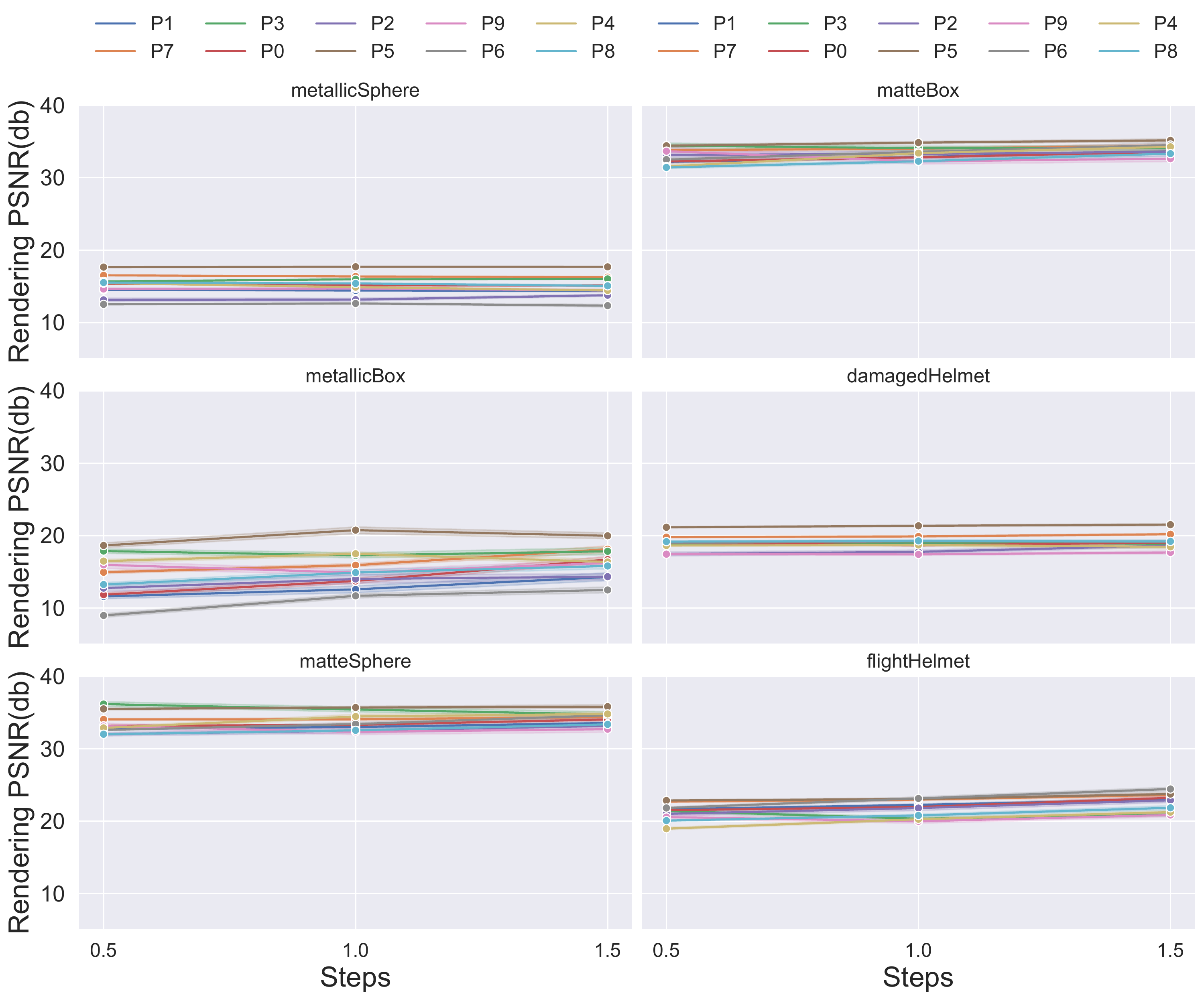}
  \caption{Detailed observation distance impact on rendering accuracy.
  }
    \label{fig:render_1_3}
\end{figure}

\begin{figure}
  \includegraphics[width=0.95\linewidth]{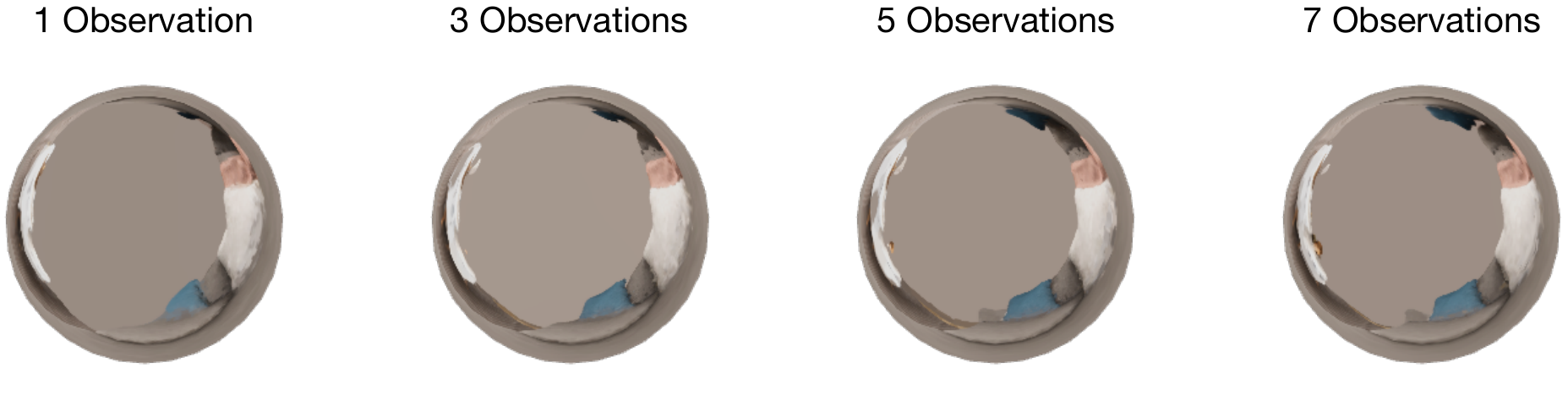}
  \caption{Rendering visual comparison for different number of near-field observations.
  }
  \label{fig:render_1_1}

\end{figure}

